\documentclass[12pt]{article}

\usepackage{amsmath,amsthm, amsfonts, amssymb, amsxtra, amsopn}
\usepackage{pgfplots}
\usepgfplotslibrary{colorbrewer}
\pgfplotsset{compat = 1.15, 
			 cycle list/Set1-3} 
\usetikzlibrary{pgfplots.statistics, pgfplots.colorbrewer} 
\usepackage{pgfplotstable}
\usepackage{graphicx,grffile}
\usepackage{multirow}
\usepackage{booktabs}
\usepackage{tcolorbox}
\usepackage{algorithm} 
\usepackage[noend]{algpseudocode} 
\usepackage{listings}
\usepackage{cmap}
\usepackage{colortbl}
\usepackage{adjustbox}
\usepackage{epsfig}

\usepackage[tableposition=top,font=small,skip=5pt,width=\textwidth]{caption}
\usepackage{subcaption}
\usepackage{makecell}

\usepackage[explicit]{titlesec}

\usetikzlibrary{patterns}

\def\bbb#1{{\color{blue}#1}}
\def\com#1{\bbb{\texttt{/\kern-1.5pt /} #1}}

\def\tau{\mathcal{T}}

\PassOptionsToPackage{hyphens}{url}
\PassOptionsToPackage{table}{xcolor}

\usepackage{hyperref}
\hypersetup{colorlinks=true,linkcolor=black,citecolor=black,urlcolor=blue,filecolor=black}
\hypersetup{pdfpagemode=UseNone,pdfstartview=}

\definecolor{darkgreen}{rgb}{0.125,0.5,0.169}

\usepackage[shortlabels]{enumitem}
\setlist[itemize]{noitemsep, topsep=0pt}

\advance\oddsidemargin by -0.45in
\advance\textwidth by 0.9in

\advance\topmargin by -0.5in
\advance\textheight by 1.0in

\def\eref#1{(\ref{#1})}

\long\def\symbolfootnotetext[#1]#2{\begingroup%
  \def\thefootnote{\fnsymbol{footnote}}\footnotetext[#1]{#2}\endgroup}

\newcommand\dunderline[3][-1pt]{{%
      \sbox0{#3}%
      \ooalign{\copy0\cr\rule[\dimexpr#1-#2\relax]{\wd0}{#2}}}}
\def\uuu{\kern-1pt\dunderline{0.75pt}{\phantom{M}}}

\DeclareMathOperator{\shuf}{shuf}
\DeclareMathOperator{\bfshuf}{\textbf{shuf}}
\DeclareMathOperator{\AI}{AI}
\DeclareMathOperator{\Bal}{Bal}
\DeclareMathOperator{\Rec}{Rec}

\def\zz{\phantom{0}}
\def\zzc{\phantom{0,}}
\def\zm{\phantom{$-$}}

\def\un{\underline{\phantom{M}}}

\title{Robustness of AI-Art Detectors under Generator Shift}

\author{Shivank Singh Thakur\footnotemark[1]\ \ \ 
Meien Li\footnotemark[1]\ \ \ 
Mark Stamp\footnotemark[1]\,\,\footnotemark[2]} 

\begin{document}

\symbolfootnotetext[1]{Department of Computer Science, San Jose State University}
\symbolfootnotetext[2]{mark.stamp$@$sjsu.edu}

\maketitle

\abstract
Text-to-image generative models have advanced rapidly, with modern
Diffusion Transformer architectures producing images that are increasingly difficult 
to distinguish from human-created artwork. 
This development has raised significant concerns regarding copyright protection, 
misinformation, fraud, impersonation, and the authenticity of digital content.
Most AI-art detectors are trained and evaluated on the same generator
family, leaving robustness to newer architectures underexplored.
In this chapter, we analyze generator shift based on a 
Stable Diffusion 3.5 Medium (SD3.5m) 
artwork dataset spanning ten art styles through reverse prompting
of held-out human artwork samples. Five detectors are trained on
U-Net-based latent diffusion artwork and evaluated in a zero-shot
cross-generator setting on the SD3.5m dataset. Deep learning models perform strongly 
in-distribution but degrade under generator shift, misclassifying many SD3.5m images 
as human while human false positives remain low. The CLIP ViT-L/14 model performs best overall, 
while Grad-CAM analysis reveals weaker and more diffuse activation on false negatives. 
These findings highlight a generalization gap in current AI-art detectors and motivate 
the development of detectors as one component of a layered defense that remains reliable across 
rapidly evolving generative architectures. 

\bigskip


\section{Introduction}\label{chap:intro}

Text-to-image generative models have greatly improved in their ability to synthesize
high-quality artwork from natural language descriptions. Given a text prompt, these
models iteratively denoise random noise into a high-resolution image that matches the
requested content and style. As architectures, training datasets, and compute budgets
have scaled up, modern systems such as DALL\textperiodcentered E, Midjourney, and Stable Diffusion
now produce images that are often difficult to distinguish from human-created 
artwork~\cite{ramesh2021zero, rombach2022high}.
Newer versions, such as Stable Diffusion~3.5 and its variants including 
Medium, Large, and Turbo, further reduce visible artifacts
and generate more coherent textures and fine details than earlier models~\cite{esser2024scaling}.

This rapid improvement creates a security and trust problem. AI-generated art can be
used for misinformation, fraud, and impersonation, especially when there is no reliable
metadata about how the image was created. 
In many real-world scenarios (social media,
messaging apps, reposted or edited content), cryptographic signatures or provenance
tags may be missing or unreliable. This motivates continued interest in image-based detection 
as an important layer of defense for estimating whether an artwork is human-created or AI-generated.

A central challenge in this setting is generator shift. Most existing AI-image detection studies train and evaluate detectors on images from the same generator family or from closely related generators (for example, Latent Diffusion models or GANs)~\cite{goodfellow2014generative, rombach2022high, wang2020cnn}.  As a result, detectors may perform well on controlled benchmark datasets, but their reliability in real deployment is far more uncertain. The images encountered after deployment may come from newer generative models that were not included during training. This chapter considers such a setting by asking whether detectors trained on artwork from earlier U-Net-based diffusion models remain reliable when evaluated on Stable Diffusion 3.5 Medium (SD3.5m), a newer Diffusion Transformer-based model~\cite{esser2024scaling}.

We model this deployment setting as an emerging-generator threat model. We assume that the defender operates an image-based detector as a screening component at a social media platform, a news verification desk, or a content-moderation pipeline. The detector uses a fixed operating threshold and has no access to labeled images from future generators. The opposing role is a synthetic-media producer who generates images using tools that were not available during detector training. Importantly, this setting does not require a sophisticated adversary who crafts adversarial perturbations or optimizes directly against the detector. The producer can simply adopt a newer publicly available generator as stronger generative models continue to emerge and become widely accessible. This creates a dynamic similar to malware detection, where new malware families may appear after a classifier has already been deployed. The threat distribution changes whereas the defensive model remains fixed and the resulting reliability gap can silently grow over time. The main operational risk is therefore a false negative, where AI-generated content is passed as human-created. This leads to an evasion-like failure mode in which the detector appears reliable on known generators but misses a substantial fraction of synthetic images from the newer generator. The formal threat model and evaluation scope are defined in Section~\ref{sec:threat_model}.

In this chapter, we investigate detector robustness under generator shift using a prompt-aligned SD3.5m dataset. 
The dataset is constructed across~10 art styles from held-out human artwork samples via reverse prompting. 
Five deep detector backbones are considered (two ResNet models~\cite{he2016resnet}, 
EfficientNet-B0~\cite{tan2019efficientnet}, ConvNeXt-Base~\cite{liu2022convnext}, and
CLIP ViT-L/14~\cite{radford2021clip}). These models are trained on in-distribution (ID) artwork from Latent
Diffusion (LDM) and Stable Diffusion 2.1 (SD2.1), then evaluated under a
shared protocol on both ID and out-of-distribution (OOD) data. This
design makes it possible to assess overall OOD degradation while also examining how the performance 
varies across styles and source generators.

The main contributions of this work are the following.
\begin{enumerate}
    \item A prompt-aligned Stable Diffusion 3.5 Medium dataset of~10,000 images across ten art styles, 
    constructed through reverse prompting of human artwork.
    \item A systematic evaluation of five deep learning detectors under generator shift, 
    measuring both in-distribution and out-of-distribution performance.
    \item A qualitative failure analysis of the case where detection fails on the newer generator, with an
    emphasis on how such failure varies across art styles.
    \item A security-oriented analysis of generator shift as an emerging synthetic-media threat, including the implications of false-negative failures for defensive content-screening and authenticity workflows.
\end{enumerate}

The remainder of this chapter is organized as follows.
Section~\ref{chap:background} reviews background on image generation
architectures, prompt engineering, detection methodologies, and relevant
datasets. Section~\ref{chap:implementation} describes dataset construction and detector implementation. 
Section~\ref{chap:results} presents our experimental results and analysis. 
Finally, Section~\ref{chap:conclusion} concludes this study and outlines possible directions for future work.

\section{Background and Related Work}\label{chap:background}

Artificial intelligence has become a major force in visual art and image synthesis. What began
as experimental algorithmic graphics has turned into an industrial-scale capability where
generative models can produce images that are often indistinguishable from human-created
artwork. This shift has driven parallel advances in forensic methods and computational models
of aesthetics. Modern AI systems can generate, analyze, critique, and imitate artistic styles. This raises
practical questions about originality, content provenance, and even the reliability of visual
evidence. 

\subsection{Computational Aesthetics: Can AI Understand Art?}

In addition to producing pixels, modern systems can evaluate
images, describe them, and optimize toward aesthetic preferences. 
Early computer vision systems focused on low-level features like color histograms, edges and
textures. Modern Multimodal Large Language Models (MLLMs) go much further, as they can
perform structured art analysis (composition, style, mood) and often agree with human
judgments about what looks ``good'' or ``beautiful''~\cite{jiang_multimodal_2025}.

By mapping images and text into a shared representation space, models such as CLIP
(Contrastive Language-Image Pre-training) can retrieve and classify artwork by abstract
concepts like ``melancholy'' or ``sublime'' and also detect concrete objects. 
This enables aesthetic reward models that steer generation toward
human-preferred styles~\cite{jiang_multimodal_2025}.

However, such model ``understanding'' is statistical and not experiential. The model learns correlations in a
high-dimensional latent space. For example, a model can recognize Impressionist brushwork or Renaissance
composition patterns, but it does not experience emotion or meaning the way human viewers
do~\cite{cetinic_understanding_2021}. Recent work adding chain-of-thought-style reasoning to MLLMs 
narrows this gap by having the model explicitly explain its judgments, which makes it appear to be more 
like an art critic~\cite{jiang_multimodal_2025}.

\subsection{Evolution of Image Generation Architectures}

Modern AI art has gone through three main phases: early visualization methods, generative
adversarial networks (GANs) and diffusion models. Each phase improved fidelity, diversity, and
controllability, and each left different forensic traces.

\subsubsection{Pre-Generative Era: DeepDream and Style Transfer (2015-2017)}

Google's DeepDream (2015) is often cited as a starting point for modern AI art~\cite{noauthor_inceptionism_nodate}. 
DeepDream used convolutional neural networks (CNNs) trained for image classification and inverted them:
instead of predicting labels from an image, it optimized the input image to strongly activate
specific neurons or layers. This produced surreal, hallucination-like images dominated by
patterns the network had learned (e.g., dog faces, eyes). DeepDream was deterministic and
limited by the features learned for classification.

Neural Style Transfer (NST) separated content (the structure of a scene)
from style (texture and color statistics). It allowed users to apply the
style of artists such as Van Gogh or Picasso to a photograph, making
AI-generated art widely accessible as a filter. However, since NST is an
image-to-image translation method rather than a fully generative model,
it stylizes existing content instead of creating new content from
scratch~\cite{gatys_imgstyletransfer_2016}.

\subsubsection{Adversarial Era: GANs (2017-2021)}


Generative Adversarial Networks (GANs) introduced adversarial training as a
powerful framework for image generation~\cite{goodfellow2014generative}. In this
setup, a generator network creates synthetic images from latent noise, while a
discriminator tries to distinguish generated images from real images. Training
proceeds as a two-player game in which the generator improves by learning to fool
the discriminator.

Key developments include the following.
\begin{itemize}
  \item StyleGAN and BigGAN delivered high-fidelity, high-resolution images. StyleGAN in
        particular allowed control over the style at different spatial scales (coarse, medium, fine),
        enabling realistic control over attributes like pose, lighting, or facial features~\cite{brock2018large, karras2019style}.
  \item GAN-based models trained on artwork datasets such as WikiArt~\cite{noauthor_wikiart_nodate} 
  	could produce novel paintings that resembled
        established art movements. The auction of the GAN-generated 
        \emph{Portrait of Edmond de Belamy} symbolized the growing cultural 
        and commercial visibility of AI-generated art~\cite{noauthor_obvious_nodate}.
\end{itemize}

Despite their success, GANs have several limitations. One common problem is mode collapse, 
where the generator produces a limited diversity of images
that still fool the discriminator. GAN-generated images may also contain characteristic spatial or
high-frequency artifacts, including checkerboard-like patterns often associated with
upsampling operations. These artifacts can provide useful signals for forensic
detection~\cite{zhang_detecting_2019}.

\subsubsection{Diffusion Era: Midjourney, DALL\textperiodcentered E and Stable Diffusion (2021-Present)}

Diffusion models use a probabilistic iterative denoising approach. They are trained to reverse a gradual noising process. Starting from a noisy image, they learn to denoise it step by step. At
inference time, generation begins from random noise and proceeds through multiple denoising
steps.

Latent Diffusion Models (LDM) achieve a balance between
computational efficiency and high-fidelity generation by operating in a
compressed latent space rather than pixel space~\cite{yang_diffusion_2024}. 
Advantages of diffusion models include better training stability and wider coverage 
of the data distribution as compared to GANs, as well as strong controllability when conditioned 
on text embeddings (for example, from models like CLIP or T5). This enables more detailed and 
precise prompt adherence~\cite{raffel2020exploring}.

\subsection{Misuse of AI-Generated Artwork}

Generative artificial intelligence (GenAI) has become increasingly widespread, substantially lowering the technical barriers to digital art creation and making creative tools more accessible to the general public. For example, text-to-image generative models such as Stable Diffusion became publicly available through online APIs and open-source repositories in mid-to-late 2022. By 2024, these models had attracted tens of millions of users~\cite{zhou2024generative}. Many researchers have emphasized the efficiency and potential of this technology for educational and academic applications. However, alongside these benefits, the misuse of GenAI has raised growing concerns, particularly regarding the unethical use and exploitation of copyrighted materials.

The harms associated with GenAI models have not yet been effectively addressed from a regulatory perspective. This is due to a combination of policy-related challenges, including inadequate regulation of social media platforms, unresolved ethical concerns, legal uncertainty, and the rapid pace of technological development.

For example, many artists have adopted various methods to prevent their work from being incorporated into GenAI training datasets, such as applying watermarks or using other protective measures~\cite{shan2023glaze}. However, these approaches do not fully deter malicious actors who continue to develop increasingly sophisticated techniques for reproducing or imitating distinctive artistic styles.

At the same time, high-quality AI-generated artwork can be used to make disinformation more persuasive and difficult to detect. Consequently, preserving the authenticity of visual content has become increasingly challenging. This erosion of trust in digital media may undermine public confidence and make it more difficult for individuals to distinguish genuine content from fabricated information.

\subsubsection{Copyright Concerns}

With the widespread adoption of the Internet, digitization of traditional artwork, online publication, and sharing of digital art have become common practices. Digital platforms provide artists with practical ways to preserve copies of their work, promote their creations, and reach broader audiences. In sharing their work online, artists often rely on copyright protection policies and the expectation that their work will not be used without authorization.

Over many years, the collective creative output of artists has contributed to the development of vast online image collections. These collections have subsequently provided the foundation for the large datasets required to train generative AI models. However, artists often have limited opportunities to refuse or prevent the unauthorized inclusion of their work in such datasets. Moreover, AI-generated artwork can blur the boundaries of ownership and authorship, particularly because relevant legal frameworks remain incomplete or insufficiently developed~\cite{bozkurt2024genai, routray2025ethical}. This legal ambiguity poses significant risks to the protection of artists' intellectual property rights.

\subsubsection{Misinformation}

The rapid development of GenAI, particularly diffusion-based image-generation models, has enabled the fast and convincing imitation of artists' styles. It has also increased the realism and persuasiveness of fabricated identities and visual content, potentially contributing to social disruption~\cite{routray2025ethical}.

An artist's body of work and distinctive style are integral components of their artistic identity. However, these characteristics can be easily exploited through generative technologies. For example, an artist's style may be imitated to appropriate its commercial value, while the artist's identity may be misrepresented to disseminate misinformation, rumors, or harmful content. Such misuse may damage the artist's reputation and weaken public trust in visual media. More broadly, the increasing prevalence of convincing fabricated content may contribute to widespread distrust and make reliable information more difficult to identify~\cite{park2026generative}.

\subsection{Prompt Engineering and Inversion}

The emergence of GenAI systems has altered the landscape of digital art. 
Unlike the traditional image generation models, current GenAI systems are capable of producing high-quality
images that closely mimic the styles of human artists, which has introduced a critical
threat to the intellectual property rights of human artists. A particular type of threat
involves an adversary passing an artist’s original artwork through a vision-language
model to obtain a natural language description, to later feed into a GenAI model to
produce a stylistically similar image. Because no pixel is directly reproduced, such
attacks are challenging and difficult to detect. Since these attacks rely on textual descriptions 
to transfer stylistic information from an original artwork to a generative model, 
understanding the role of prompts becomes essential.

\subsubsection{Image Captioning as Prompt}

Prompt engineering is the practice of crafting text inputs that steer the model toward a desired
style, composition, or mood. This often relies on modifiers i.e., keywords that the model has
learned to associate with specific visual attributes, such as ``unreal engine'', ``octane render'',
or ``volumetric lighting''.

Image captioning models like BLIP~\cite{li2022blip} or LLaVA~\cite{liu_visual_2023} 
can generate prompts from reference images helping automate prompt creation. 
However, generic captions usually describe objects and
actions and often miss fine-grained stylistic details (brushwork, camera model, rendering engine) 
needed to reproduce an artwork's look.

\subsubsection{Reverse Prompt Engineering and Prompt Inversion}

Reverse prompt engineering (RPE) aims to reconstruct a prompt that could possibly have
generated a given image. Common methods include the following.
\begin{itemize}
  \item PEZ (Hard Prompts made EaZy) uses gradient-based optimization on discrete
        tokens to minimize CLIP loss between the prompt and target image. It often discovers
        trigger tokens that work numerically but do not have a human-readable meaning 
        (for example, ``\texttt{horse cat 8k!!}'')~\cite{wen_hard_2023}.
  \item CLIP Interrogator combines a captioning model with a large database of style tags,
        then searches for tags that maximize CLIP similarity. This produces human-readable
        prompts like ``\texttt{digital art by [Artist], trending on ArtStation}'' but performance is limited by
        the tag library~\cite{pharmapsychotic_pharmapsychoticclip-interrogator_2023}.
  \item ARPO (Automatic Reverse Prompt Optimization) iteratively generates images from
        candidate prompts, compares them to the target and updates the prompt. It can find
        high-fidelity prompts that are easy to edit, but is computationally expensive because it
        requires many generation steps~\cite{ren_reverse_2025}.
\end{itemize}

These methods converge on prompts that are semantically close to the original, effectively reconstructing 
a textual specification of the generation process that can be used for attribution, reproducibility or further analysis.

\subsection{Methodologies for Detecting AI-Generated Artwork}

As generative models have improved, detection has moved from obvious visual glitches to more
subtle statistical, semantic, and model-based cues. Current methods can be grouped into
artifact analysis, generative fingerprinting and semantic reasoning.

\subsubsection{Artifact Analysis: Frequency and Spatial Domains}

Frequency-domain analysis uses techniques like the Fast Fourier Transform (FFT) to study the
spectrum of an image. GANs with transposed convolutions can leave periodic patterns that
show up as spikes or checkerboard structures in the frequency domain. Similar but more subtle
artifacts can sometimes be found in diffusion images due to their deconvolution modules~\cite{convertini_discrete_2024}.

Spatial-domain analysis focuses directly on pixel patterns, such as texture inconsistencies or
unusual local noise structures. These signals are fragile and common post-processing operations
(like JPEG compression, blurring and resizing) can weaken or remove them, which makes pure
artifact-based detectors vulnerable~\cite{cai_robust_2025}.

\subsubsection{Generative Fingerprinting: Diffusion Reconstruction Error}

DIRE (Diffusion Reconstruction Error) uses a diffusion model as a detector instead of only as a generator~\cite{wang_dire_2023}. The method assumes that if an image was created by a given diffusion model, it lies close to that model’s image manifold. In practice, the image is first inverted into a noisy latent using the same model, then reconstructed by running the diffusion process forward and finally an error is computed between the original and reconstructed images. Synthetic images that match the model tend to reconstruct with low error, while real images or out-of-distribution images generally reconstruct with higher error because important details are lost in the inversion–reconstruction cycle. DIRE is effective even for unseen diffusion models because it exploits properties of the
diffusion process itself rather than specific visual artifacts. Its main drawback is computational
cost due to repeated denoising steps.

\subsubsection{Human Perception and Semantics}

Human judgment is still important even with strong automatic detectors. Recent work that compares
human and machine performance on real versus AI-generated images shows that people are unreliable
at this task, as they tend to rely more on semantic and physical cues than on low-level artifacts.
Typical signals include lighting and shadows (e.g., inconsistent shadow directions, incorrect
reflections, or specular highlights that do not match the scene's light sources) and anatomy and
geometry (e.g., subtle deformations in hands, faces, or object boundaries, or physically impossible
arrangements such as floating objects)~\cite{kamali_characterizing_2025}. Benchmarks such as AnomReason~\cite{tan_semantic_2025} explicitly test whether models
can detect these kinds of semantic and physics-based anomalies rather than only exploiting texture
statistics.

Empirical studies report that human participants misclassify a large fraction of images (real versus AI),
while the best automatic models achieve substantially lower error rates on clean, in-distribution data~\cite{ha_organic_2024}.
However, automatic detectors can be fragile under adversarial perturbations, meaning small and carefully designed
changes to pixels can drastically reduce their accuracy, even when the images still look the same
to humans. For artwork, large-scale studies with crowdworkers and expert artists show a similar pattern. Experts tend to outperform non-experts and are better at noticing semantic and physical
inconsistencies, but they also exhibit a ``skepticism bias'', i.e., they label unusual or abstract
human art as AI-generated, which can lead to a high false-positive rate~\cite{ha_organic_2024}.

Overall, these results suggest that humans and models make different kinds of errors. Humans are more
robust to small pixel changes and better at reasoning about semantics and physical errors, but
are biased and overly-suspicious especially for atypical styles. Models are more consistent and sensitive
to subtle statistical patterns, but can be fooled by low-level perturbations and shifts in distribution 
and often struggle with semantic anomaly detection.

\subsection{Deep Learning for Detection}\label{sec:dl_architectures}

Deep learning architectures for visual recognition have evolved rapidly,
progressing from early convolutional networks through residual learning,
efficient scaling, and modernized CNN designs, to transformer-based
models that capture global spatial dependencies. This section reviews
the key architectural developments most relevant to image classification
and representation learning. Table~\ref{tab:backbone_summary} provides 
a summary comparison of the models used in this study.

\begin{table}[!htb]
\centering
\caption{Summary of deep learning backbone architectures}\label{tab:backbone_summary}
\begin{adjustbox}{scale=0.785}
\begin{tabular}{llccl}
\toprule
\textbf{Model} &
  \textbf{Architecture} &
  \textbf{Parameters} &
  \textbf{Year} &
  \textbf{Pretraining} \\
\midrule
ResNet-18~\cite{he2016resnet}
  & Residual CNN          & \zz11.7M  & 2016 & ImageNet-1K \\
ResNet-50~\cite{he2016resnet}
  & Residual CNN          & \zz25.6M  & 2016 & ImageNet-1K \\
EfficientNet-B0~\cite{tan2019efficientnet}
  & Compound-scaled CNN   &  \zz\zz5.3M  & 2019 & ImageNet-1K \\
  CLIP ViT-L/14~\cite{radford2021clip}
  & Vision Transformer    & 307.4M & 2021 & 400M image-text pairs \\
ConvNeXt-Base~\cite{liu2022convnext}
  & Modernized CNN        & \zz88.6M  & 2022 & ImageNet-1K \\
\bottomrule
\end{tabular}
\end{adjustbox}
\end{table}

\subsubsection{Residual Networks}\label{sec:resnet}

Training very deep convolutional networks is hindered by the degradation
problem, where adding more layers increases training error,
suggesting an optimization difficulty rather than overfitting. Residual
learning addresses this by introducing skip connections that add the
block input directly to its output, so the network learns residual
mappings rather than full unreferenced mappings, allowing gradients to
propagate effectively through very deep architectures.
The resulting ResNet family, spanning 18 to over 150 layers, achieved 
state-of-the-art results on ImageNet and became a foundational
backbone architecture widely adopted across detection, segmentation, and
representation learning tasks~\cite{he2016resnet, russakovsky2015imagenet}.

\subsubsection{EfficientNet}\label{sec:efficientnet}

Scaling network depth, width, and input resolution independently yields
suboptimal results, as these dimensions are interdependent. Compound
scaling addresses this by uniformly scaling all three dimensions
simultaneously using a fixed coefficient derived from a constrained grid
search, producing models that achieve significantly better accuracy while using 
fewer parameters and less computation than many earlier models. 
The EfficientNet family showed that scaling strategy is just as important as 
choosing the model design itself~\cite{tan2019efficientnet}.

\subsubsection{Vision Transformers and CLIP}\label{sec:vit_clip}

Transformer architectures were adapted to image recognition by
splitting images into fixed-size patches, projecting each patch into a
token embedding, and applying multi-head self-attention across the full
sequence~\cite{dosovitskiy2021vit}. Unlike convolutional networks, this
approach captures global spatial relationships from the first layer
without locality constraints, and achieves strong performance when
pretrained on sufficiently large datasets. This was extended further by
training a Vision Transformer (ViT) through contrastive learning on~400 million
image-text pairs, aligning visual and textual representations in a shared
embedding space~\cite{radford2021clip}. This Contrastive Language-Image
Pretraining (CLIP) objective produces semantic, transferable
representations that generalize broadly across tasks without
task-specific fine-tuning, and has become one of the most widely used
backbones for zero-shot and transfer learning applications.

\subsubsection{ConvNeXt}\label{sec:convnext}

As ViT gained traction, the convolutional design space
was revisited by systematically applying transformer-inspired choices 
(larger depthwise kernels, inverted bottleneck blocks, layer normalization,
and GELU activations~\cite{hendrycks2016gaussian}) to a standard ResNet backbone. 
The resulting ConvNeXt family matches ViT performance on
standard benchmarks while retaining the computational efficiency of
convolutional architectures, demonstrating that the CNN-transformer
performance gap was largely attributable to training procedures and
design choices rather than a fundamental limitation of convolutions~\cite{liu2022convnext}.

\subsection{Public Datasets for AI Artwork Detection}

A variety of public datasets are available for AI-image and AI-artwork detection covering
different domains, generators, and threat models. Some are large, general-purpose benchmarks
designed to learn broad, model-agnostic forensic features, while others focus specifically on
artistic styles or robustness to real-world degradations and adversarial perturbations.

Table~\ref{tab:datasets} summarizes key properties of representative datasets, including total
size, real/fake composition, real-image sources, underlying generators,
and primary use cases. Robust AI-artwork detectors are usually trained and
evaluated using a combination of resources, including large general datasets for broad coverage,
art-focused datasets for style awareness, and robustness-oriented datasets to stress-test
models under realistic noise, compression, and attack settings.

\begin{table}[!htb]
  \centering
  \caption{Summary of public AI-generated image and art detection datasets}\label{tab:datasets}
  \begin{adjustbox}{scale=0.775}
    \begin{tabular}{lccccc}
      \toprule
      \multirow{2}{*}{\textbf{Dataset}} &
      \textbf{Total} &
      \textbf{Real/Fake} &
      \textbf{Real} &
      \multirow{2}{*}{\textbf{Generators}} &
      \textbf{Primary} \\
      &
      \textbf{images} &
      \textbf{split} &
      \textbf{sources} &
      &
      \textbf{use case} \\
      \midrule
      \multirow{2}{*}{ArtiFact~\cite{rahman_artifact_2023}} 
      & \multirow{2}{*}{$\sim$2.5M}
      &
        \multirow{2}{*}{Balanced} &
        COCO, LSUN, &
        GANs,  &
        Social-media \\
        &
        &
        &
        FFHQ, AFHQ 
        &
        diffusion
        &
        robustness \\
      \midrule
      \multirow{2}{*}{GenImage~\cite{zhu2023genimage}} 
      &
      \multirow{2}{*}{$\sim$2.7M} 
      &
      \multirow{2}{*}{ImageNet} 
      &
      \multirow{2}{*}{1000 classes} 
      &
      Midjourney, SD, 
      &
        Forensics, \\
        &
        &
        &
        &
        GLIDE, etc. &
        cross-generation \\
      \midrule
      \multirow{2}{*}{ArtBench-10~\cite{liao_artbench_2022}} &
        \multirow{2}{*}{60k} &
        \multirow{2}{*}{All real} &
        WikiArt &
        \multirow{2}{*}{---} &
        Style-aware \\
        &
        &
        &
        (10 styles)
        &
        &
        benchmarks \\
      \midrule
      \multirow{2}{*}{AI-ArtBench~\cite{silva_artbrain_2024}} &
        \multirow{2}{*}{$\sim$185k} &
        60k real &
        ArtBench-10 &
        Latent Diffusion,&
        Human vs.\\
        &
        &
        125k fake 
        &
        (real)
        &
        SD &
         AI art styles \\
      \midrule
      \multirow{2}{*}{Human-Art~\cite{ju_human-art_2023}} &
        \multirow{2}{*}{50k} &
        Mixed real &
        Natural photos &
        \multirow{2}{*}{Various} &
        Detection \\
        &
        &  
        and stylized
        &
        plus WikiArt &
        &
        across styles \\
      \midrule
      \multirow{2}{*}{CIFAKE~\cite{bird_cifake_2024}} &
        \multirow{2}{*}{120k} &
        60k real &
        \multirow{2}{*}{CIFAR-10} &
        \multirow{2}{*}{SD v1.4} &
        Edge \\
        &
        &
        60k fake &
        &
        &
        benchmarking \\
      \midrule
      \multirow{2}{*}{COCO-Fake~\cite{amoroso_parents_2024}} &
        \multirow{2}{*}{$\sim$1.2M} &
        \multirow{2}{*}{Paired fakes} &
        \multirow{2}{*}{MS-COCO} &
        \multirow{2}{*}{SD v1.4, v2.0} &
        Caption--image \\
        &
        &
        &
        &
        &
        consistency \\
      \midrule
      \multirow{2}{*}{RAID~\cite{eddoubi_raid_2025}} &
        \multirow{2}{*}{$\sim$72k} &
        \multirow{2}{*}{Balanced} &
        Subset of &
        \multirow{2}{*}{SD variants} &
        Adversarial \\
        &
        &
        &
        LAION-400M
        &
        &
        robustness \\
      \bottomrule
    \end{tabular}
  \end{adjustbox}
\end{table}

Recent research shows a shift from GAN-based image generation to modern 
diffusion and transformer-based systems. Detection methods
have expanded from low-level artifact analysis to feature-based and
semantically informed deep learning approaches. Artwork-specific detection studies 
have also explored human-versus-AI classification using classical machine learning and 
deep learning models, showing that strong performance is possible under controlled 
style and generator settings~\cite{li_detecting_2025}. However, current
benchmarks still provide limited coverage of prompt-aligned, art-specific
generator shift, especially for newer models such as Stable Diffusion~3.5. 
This gap motivates the present study, which evaluates five deep backbone detectors 
under a common experimental protocol and tests their ability to generalize from earlier 
generators to a new prompt-aligned Stable Diffusion~3.5 Medium dataset.

\section{Dataset and Implementation}\label{chap:implementation}

This section describes the construction of the datasets used for training and out-of-distribution
(OOD) evaluation, and the design and training of the binary detectors that we evaluate 
as defensive screening components.
We also describe our detector architecture, training procedure,
threshold selection strategy, and evaluation protocol.

\subsection{Threat Model and Evaluation Scope}\label{sec:threat_model}

We consider a post-deployment generator-shift scenario in which an image-based detector is used as a defensive screening component to distinguish human-created artwork from AI-generated artwork. The defender may represent a content-moderation platform, media-verification service, digital-art marketplace, or another organization that reviews the claimed origin of visual content. We assume that the detector has been trained using images from known generator families, deployed with a fixed decision threshold, and has no access to labeled samples from generators that emerge after deployment. 

The opposing role is a synthetic-media producer who uses a newer, publicly available image generator that was not represented during detector training. The producer is not assumed to have access to the detector's parameters, gradients, training data, or internal confidence scores, and does not construct adversarial pixel perturbations. Instead, the relevant evasion mechanism is the adoption of a generator whose architecture and learned visual characteristics differ from those represented in the training data. The primary failure of interest is therefore a false negative, in which an AI-generated image is classified as human-created to pass the screening mechanism and evade the detector. False positives are also important because incorrectly labeling human-created artwork as AI-generated can cause attribution errors and reputational harm to artists, while also undermining public trust in detection systems. 

This threat model is operationalized by training detectors on artwork generated by Latent Diffusion and Stable Diffusion 2.1 and evaluating them, without retraining or threshold adjustment, on artwork generated by Stable Diffusion 3.5 Medium. The SD3.5m images are produced from prompts derived from held-out human artwork so that semantic content and artistic-style coverage remain comparable while the underlying generator changes. The scope of this study is limited to image-level human-versus-AI artwork detection. It does not directly evaluate metadata manipulation, watermark removal, account compromise, adversarial perturbations or the intent of the content producer. Detector predictions should therefore be interpreted as screening signals for further verification rather than as definitive proof of authenticity or malicious use.

\subsection{Experimental Setup}\label{sec:hw_sw}

Table~\ref{tab:hw_config} lists the hardware configuration used for the experiments. 
All deep learning workloads used PyTorch and
torchvision, with CUDA 12.8 as the compute backend~\cite{paszke2019pytorch}. 
Image generation was performed with the HuggingFace diffusers
library, and pretrained CLIP and BLIP model weights
are loaded via HuggingFace transformers~\cite{wolf2020huggingface}.

\begin{table}[!htb]
\centering
\caption{Hardware configuration used for all experiments}\label{tab:hw_config}
\begin{adjustbox}{scale=0.85}
\begin{tabular}{ll}
\toprule
\textbf{Component} & \textbf{Details} \\
\midrule
Compute platform & Google Colab Pro \\
GPU              & NVIDIA A100-SXM4-80GB \\
CPU              & Intel(R) Xeon(R) CPU @ 2.20GHz \\
System RAM       & 167 GB (high-memory instance) \\
Storage          & Google Drive, Colab Disk \\
\bottomrule
\end{tabular}
\end{adjustbox}
\end{table}
 
All large image directories are archived as \texttt{zip} files on Google Drive and extracted to
the Colab instance's local disk at the start of each session using the zipfile module. 
This reduced dataset loading time from several hours of cloud-based file access to a 
single bulk extraction step completed in only a few minutes.

\subsection{Artwork ID Dataset}\label{sec:31_artwork_dataset}

AI-ArtBench is used as the primary in-distribution (ID) dataset in this study~\cite{silva_artbrain_2024}. It is an art-focused benchmark for detecting and attributing AI-generated artwork, containing approximately~185,000 images across~10 artistic styles: Art Nouveau, Baroque, Expressionism, Impressionism, Post-Impressionism, Realism,
Renaissance, Romanticism, Surrealism, and Ukiyo-e. It includes~60,000 human artwork samples and~125,015 AI-generated images produced by Latent Diffusion (LDM) and Stable Diffusion 2.1 (SD2.1) models. Its human subset is derived from ArtBench-10, a standardized and class-balanced dataset of~60,000 artwork images across the ten styles~\cite{liao_artbench_2022}.

AI-ArtBench includes three source categories namely human, LDM, and SD2.1. In this chapter, we consider a binary human-versus-AI setting by merging the two synthetic classes. The final in-distribution splits used in the experiments consist of~105,000 training samples, 15,000 validation samples, and 30,000 test samples. This benchmark provides the core data foundation for training and evaluating the detector in a style-diverse artwork setting.

\begin{figure}[!htb]
  \centering
  \includegraphics[width=0.85\textwidth]{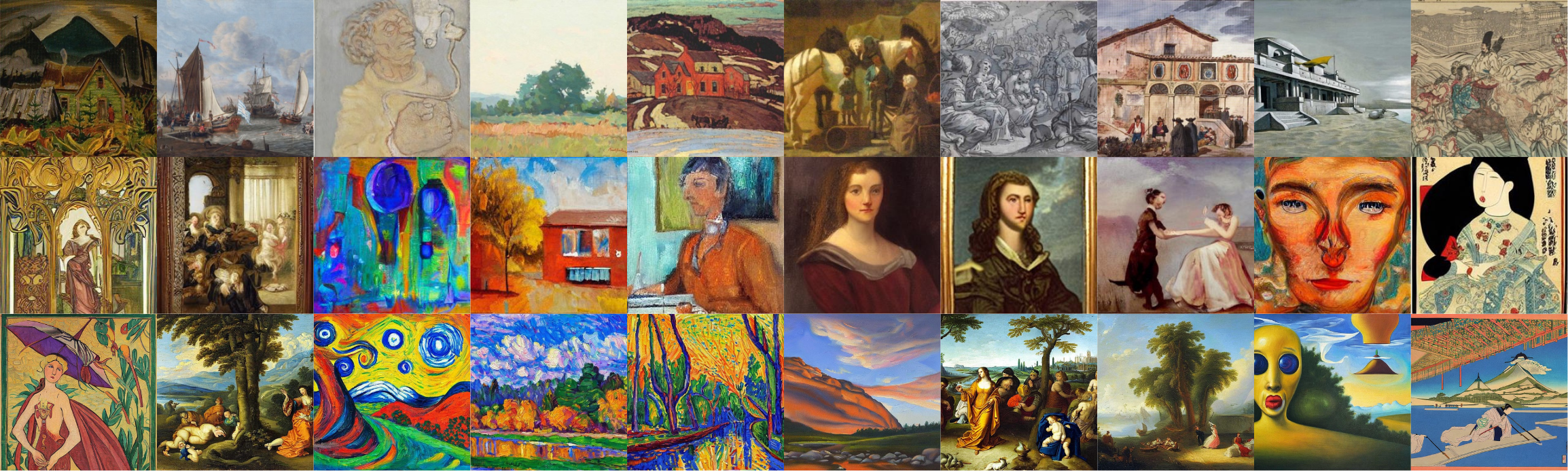}
  \caption{Samples from the AI-ArtBench dataset: human artwork (top row), LDM-generated artwork (middle row), and SD2.1-generated artwork (bottom row) across 10 art styles, shown left to right: Art Nouveau, Baroque, Expressionism, Impressionism, Post-Impressionism, Realism, Renaissance, Romanticism, Surrealism, and Ukiyo-e}
  \label{fig:human_ld_examples}
\end{figure}

\subsection{Stable Diffusion OOD Dataset}\label{sec:32_ood_dataset}

The main challenge that we address is whether a detector trained on images from LDM and SD2.1 can 
remain reliable when an adversary switches to a newer and architecturally different generator as defined in Section \ref{sec:threat_model}. To study this, 
a dataset of~10,000 images was constructed using Stable Diffusion 3.5 Medium (SD3.5m)~\cite{esser2024scaling}. SD3.5m was selected because it is a newer member of the Stable Diffusion family which Stability~AI describes as offering improved prompt adherence, image quality, and practical use on accessible hardware. Unlike the training U-Net generators, SD3.5m is built on the Diffusion Transformer (DiT) architecture rather than a convolutional U-Net. DiT supports richer text-to-image conditioning through a multi-encoder text pipeline ~\cite{peebles2023dit}. These architectural differences make SD3.5m a challenging and realistic test of cross-generator generalization.

In this study, we use the dataset for out-of-distribution evaluation. All SD3.5m images are generated using prompts derived from human artwork samples through a reverse prompting pipeline. This helps make the OOD images semantically comparable to the human artwork samples and reduces content mismatch. Essentially this prompt-aligned construction means that any decrease in detector performance is more likely to be caused by the change in generator rather than by differences in image content or style distribution.

\begin{figure}[!htb]
\centering
\includegraphics[width=0.85\textwidth]{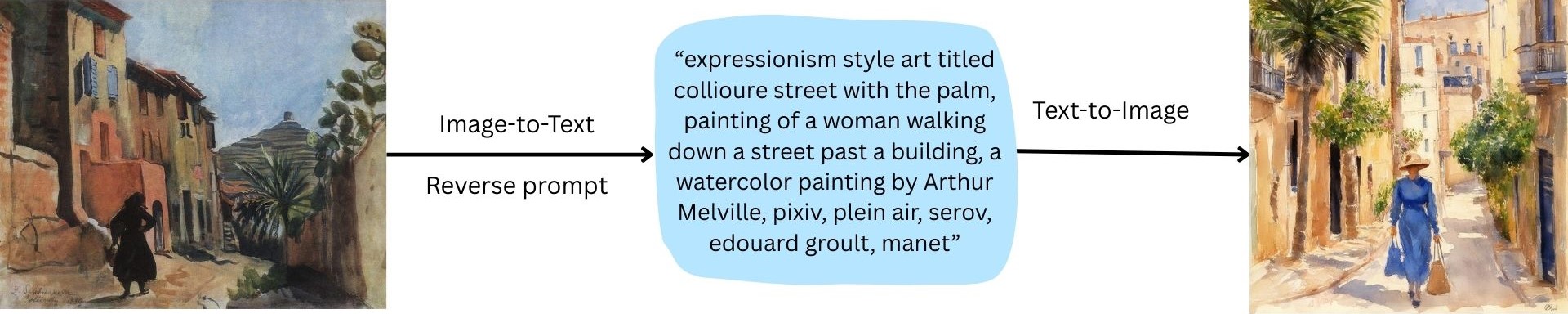}
\caption{Overview of the prompt-aligned SD3.5m dataset construction pipeline}\label{fig:ood_pipeline}
\end{figure}

The pipeline illustrated in Figure~\ref{fig:ood_pipeline} consists of four
stages: human artwork selection, reverse prompting via CLIP Interrogator, prompt
augmentation and controlled image generation with SD3.5m.

\subsubsection{Held-Out Human Reference Set}

The source images for the reverse prompting pipeline are a separate set of~10,000 human artwork samples selected from the broader artwork corpus and reserved as a held-out subset. These images were sampled across the~10 art styles used in this study, with~1,000 artwork samples per style, so that the extension would preserve stylistic coverage consistent with the rest of the dataset. The held-out artwork samples were excluded from the in-distribution training, validation, and test splits to prevent data leakage. 
These images serve as input to
the reverse prompting pipeline and as the human-class reference set in the final OOD test set.

\subsubsection{Image-to-Text via Image Captioning}\label{sec:clip_interrogator}

We used CLIP Interrogator to generate content-aware and style-consistent text prompts from human artwork~\cite{pharmapsychotic_pharmapsychoticclip-interrogator_2023}. This is a reverse prompt 
engineering tool that combines a visual
captioning model with a large database of style, artist, and aesthetic tags ranked
by CLIP similarity to the input image ~\cite{radford2021clip}. 

The system was configured with the BLIP-Large captioning model
and the CLIP ViT-L/14 vision encoder~\cite{li2022blip}. The chunk size and flavor intermediate
count were both set to 2{,}048, which is the recommended value for the ViT-L/14
backbone to accommodate its higher-dimensional embedding space.
In classic mode, CLIP Interrogator first generates a BLIP caption describing
the image content, then appends the highest-scoring style, artist, and flavor tags
from its curated databases. This produces human-readable prompts such as
%
%

\bigskip

\texttt{a painting of a family with a bird on their shoulder, a painting by Jacob}

\texttt{Jordaens, baroque, oil on canvas, Flemish baroque, Dutch golden age}

\bigskip

%
\noindent Each artwork image was processed individually, 
producing one prompt per image and saved to a CSV file along with the source filename.
The CLIP Interrogator runs on the NVIDIA A100 GPU (see Section~\ref{sec:hw_sw}) 
at approximately~1.5 images per second.


\subsubsection{Prompt Augmentation}

A prompt augmentation step is applied to captions before image generation to ensure the SD3.5m output reflects the target art style. The final prompt for each artwork is made by combining three components namely the target art style, the painting title parsed from the source image filename and the CLIP-based caption from the previous stage. This enriches the caption with explicit style information while preserving the semantic content extracted by CLIP Interrogator. The prompt template used for augmentation is
%
$$
\mbox{\texttt{\{art\un style\} style art titled \{painting\un name\}, \{CLIP\un caption\}}}
$$


\subsubsection{Text-to-Image Generation with Stable Diffusion}

For our OOD dataset, image generation was performed using the HuggingFace \texttt{diffusers} library with the
\texttt{stabilityai/stable-diffusion-3.5-medium} checkpoint. Images were generated at a fixed resolution of $768 \times 768$ pixels using 28
denoising steps and a classifier-free guidance scale of 4.5. This configuration provides good adherence to
the prompts without overly constraining the output. The \texttt{FlowMatchEulerDiscreteScheduler} is used as recommended for SD3.x models with
bfloat16 precision to balance image quality and memory usage~\cite{esser2024scaling}.  Memory-efficient attention (xFormers) was enabled to lower memory usage and support faster image generation (see Section~\ref{sec:hw_sw} for environment specification). 
These configuration parameters are summarized
in Table~\ref{tab:sd35m_config}. 

\begin{table}[!htb]
\centering
\caption{Stable Diffusion~3.5 Medium generation configuration used to construct
the OOD test set}\label{tab:sd35m_config}
\begin{adjustbox}{scale=0.85}
\begin{tabular}{ll}
\toprule
\textbf{Parameter}             & \textbf{Value} \\
\midrule
Model checkpoint               & \texttt{stabilityai/stable-diffusion-3.5-medium} \\
Output resolution              & 768 $\times$ 768 pixels \\
Number of denoising steps      & 28 \\
Classifier-free guidance scale & 4.5 \\
Scheduler                      & \texttt{FlowMatchEulerDiscreteScheduler} \\
Precision                      & bfloat16 \\
Seeding strategy               & SHA-256 hash of source filename \\
Average generation time        & 3.2 seconds per image \\
Average GPU utilization        & 86\% \\
Average GPU power              & 333 W \\
Average GPU energy per image   & 0.29 Wh \\
\bottomrule
\end{tabular}
\end{adjustbox}
\end{table}

The randomness is controlled
through a combination of a global seed and a per-image seed derived deterministically
from the input filename using a SHA-256 hash, ensuring that each human artwork always maps
to the same generated image. Output files are named to encode style, subject, and seed, for example
$$
    \mbox{\texttt{ai-sd35m-\textless style\textgreater\un \textless painting\textgreater\un \textless seed\textgreater.jpg}}
$$
An audit log is saved for every image and includes the original image filename, injected style,
parsed painting title, final prompt, seed, configuration parameters, output path, runtime,
timestamp and status message. GPU telemetry (utilization, memory, power, temperature) was sampled at 500\,ms
intervals during each generation call using NVIDIA NVML for resource monitoring.

\begin{figure}[!htb]
  \centering
  \includegraphics[width=0.85\textwidth]{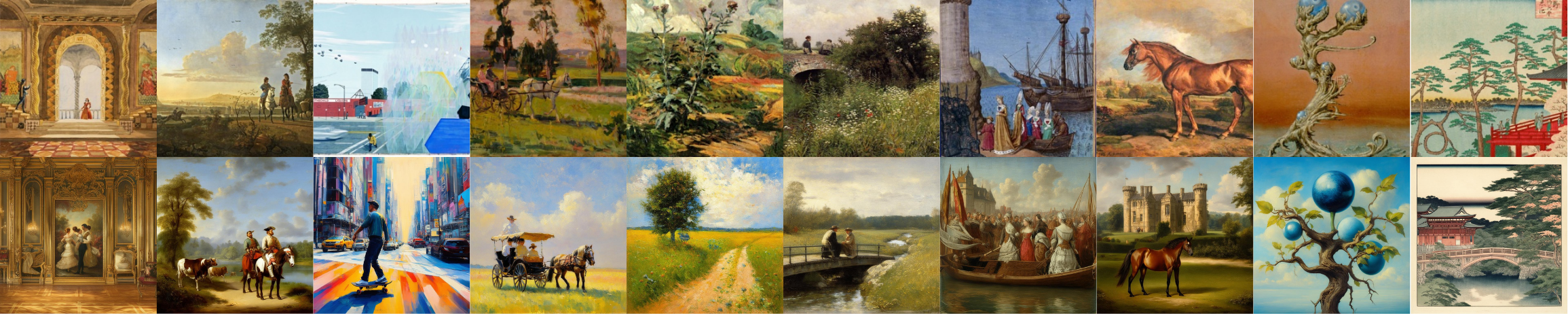}
  \caption{Samples of human artwork (top row) and prompt-aligned SD3.5m-generated artwork samples (bottom row) across~10 art styles, shown left to right: Art Nouveau, Baroque, Expressionism, Impressionism, Post-Impressionism, Realism, Renaissance, Romanticism, Surrealism, and Ukiyo-e}
  \label{fig:sd35m_examples}
\end{figure}

The complete SD3.5m AI dataset contains~10,000
images consisting of~1,000 prompt-aligned SD3.5m images per style. Sample generated images alongside their source
human artwork samples are shown in Figure~\ref{fig:sd35m_examples}. This dataset can test detector behavior under
a realistic change in the generative model, and the pipeline is scalable for future experiments.

\subsection{Dataset Quality Evaluation}\label{sec:33_dataset_quality}

The quality of the SD3.5m dataset extension was evaluated using a set of complementary metrics covering prompt alignment, distributional similarity, duplicate detection and visual diversity. A subset of the data was used for the evaluation 
which consists of~5,000 generated
images (500 per style) and~5,000 corresponding human artwork samples. Four metrics are computed: CLIPScore~\cite{hessel_clipscore_2021},
Fr\'{e}chet Inception Distance (FID)~\cite{heusel_gans_2017},
Kernel Inception Distance (KID)~\cite{binkowski_demystifying_2018}
and Learned Perceptual Image Patch Similarity (LPIPS) perceptual diversity~\cite{zhang_unreasonable_2018}.

CLIPScore measures how well each generated image matches its generation
prompt. All images and their corresponding prompts were passed through the CLIP
ViT-L/14 model and CLIPScore was computed as the cosine similarity between the
projected image and text embeddings. A shuffled-prompt baseline
was also computed by randomly reassigning prompts to images (five repetitions),
providing a lower bound that reflects what CLIP similarity looks like when
image-prompt pairing is broken.

FID and KID measure the statistical distance between the distribution of
generated images and the distribution of real human artwork samples for the same style,
both computed in Inception feature space at~$256\times 256$ resolution.
A human-versus-human (HvH) baseline was also computed by repeatedly splitting
the human set into two random halves and measuring FID/KID between those halves,
providing a lower bound that reflects the natural within-style variation among
human artwork samples.

LPIPS diversity measures the average perceptual distance between randomly
sampled pairs of generated images within the same style, using an AlexNet backbone.
Higher values indicate that the generated images are visually diverse rather than
collapsing into a small set of visual patterns. We compute the LPIPS diversity score for the generated dataset as the mean perceptual distance across~1,000 randomly sampled intra-style image pairs.

Duplicate analysis uses perceptual hashing (pHash) to detect duplicate candidates with identical pHash values and near-duplicates.
For this metric, we consider Hamming distance less than or equal to~4.


\begin{table}[!htb]
\centering
\caption{Dataset quality metrics for the SD3.5m OOD dataset (5,000 images,
500 per style; CLIP and $\mbox{CLIP}_{\shuf}$ shown in units of $\times10^{-2}$;
KID and HvH KID shown in units of $\times10^{-4}$)}
\label{tab:dataset_quality}
\begin{adjustbox}{scale=0.785}
\begin{tabular}{lccccccc}
\toprule
\textbf{Style}
  & \textbf{CLIP}
  & \textbf{$\mbox{CLIP}_{\bfshuf}$}
  & \textbf{FID}
  & \textbf{HvH FID}
  & \textbf{KID}
  & \textbf{HvH KID}
  & \textbf{LPIPS} \\
\midrule
Art Nouveau        & 28.52 & 11.70 & 119.67 & 126.21 & 214.0 & $-$0.46 & 0.712 \\
Baroque            & 28.14 & 12.87 & 110.10 & 116.30 & 226.0 &    \zm0.70 & 0.660 \\
Expressionism      & 27.32 &  \zz9.33 & 130.60 & 137.77 & 215.0 &    \zm1.92 & 0.740 \\
Impressionism      & 27.91 & 10.19 & 120.43 & 126.47 & 256.0 & $-$3.72 & 0.712 \\
Post-Impressionism & 27.63 &  \zz9.73 & 124.95 & 124.06 & 266.0 & $-$0.66 & 0.725 \\
Realism            & 28.81 & 10.31 & 118.20 & 130.44 & 219.0 & $-$3.98 & 0.706 \\
Renaissance        & 26.37 & 12.98 & 104.76 & 109.60 & 213.0 & $-$2.63 & 0.673 \\
Romanticism        & 28.49 & 10.79 & 103.66 & 118.24 & 175.0 &    \zm0.17 & 0.690 \\
Surrealism         & 28.82 & 11.86 & 128.19 & 148.35 & 187.0 & $-$2.16 & 0.736 \\
Ukiyo-e            & 30.93 & 22.55 & 139.37 &  \zz83.14 & 674.0 & $-$3.67 & 0.681 \\
\midrule
\textbf{Overall}   & \textbf{28.29} & \zz\textbf{7.53} & \zz\textbf{43.84}
                   & \zz\textbf{24.16} & \textbf{180.0} & \zm\textbf{0.38} & \textbf{0.724} \\
\bottomrule
\end{tabular}
\end{adjustbox}
\end{table}

Table~\ref{tab:dataset_quality} reports the per-style and overall results. The overall FID, KID and shuffled-prompt CLIP baseline are computed on the pooled multi-style dataset and not as averages of the per-style rows. For the pooled dataset spanning all ten styles, the reported overall CLIPScore is~28.29, compared to a shuffled-prompt baseline of~7.53, yielding a margin of~20.76 in the reported table units. This large gap
suggests that the generated images remain meaningfully aligned with their source
prompts. The overall FID is~43.84, compared to a human-versus-human baseline FID
of~24.16, indicating that the SD3.5m images remain distributionally distinct from
the human reference set. The
reported overall KID is~180.0, corresponding to a raw KID value of~0.0180, whereas
the corresponding human-versus-human KID baseline is~0.38 in the table units and
therefore remains near zero. The trend is consistent with the FID scores and supports distributional separation when measured under the second metric. The average diversity by the LPIPS criterion is~0.72 and the style-wise LPIPS scores are all in the mid to high ranges, suggesting a good amount of visual diversity across the images. There were no duplicate images found by the pHash metric and no near duplicate pairs within each style.

Ukiyo-e has the highest raw CLIPScore (30.93), but also a high shuffled baseline (22.55). This suggests the score partly reflects shared style features, not only exact image–prompt alignment. Ukiyo-e also has the largest FID score (139.37) and KID (674.0), making it a clear outlier. By comparison, Romanticism (103.66) and Renaissance (104.76) are associated with the smallest FID values among the ten styles.

In general, we see that the SD3.5m dataset retains the style coverage and prompt relevance while remaining different from the human reference dataset. This makes it suitable for style-aware out-of-distribution evaluation under generator shift.

\subsection{Detector Implementation}\label{sec:detectors}

Image detection is organized as a modular framework for evaluating multiple deep learning models on the task of binary human-versus-AI artwork detection. We consider several detector families, including ResNet, EfficientNet, ConvNeXt, and CLIP-ViT, all studied under a shared in-distribution and out-of-distribution evaluation protocol.

\subsubsection{Architecture Overview}\label{sec:arch_overview}

All detectors follow a frozen-backbone, linear-probe design. A pretrained deep
network serves as a fixed feature extractor, the final classification layer is
removed and the penultimate feature vector is extracted for every input image.
A lightweight linear classification head is then trained on top of these frozen features to perform binary
human-versus-AI classification. This approach has three main benefits. First, it
reduces the risk of catastrophically fine-tuning a large pretrained model on a
relatively small dataset. Second, feature extraction can be precomputed once
for the entire dataset, making the head-training loop extremely fast. Third,
the frozen representations provide a controlled comparison across backbone families
without confounding effects from differences in fine-tuning schedules. 
This also models a realistic deployment scenario where the OOD performance reflects cross-generator transfer without adaptation to SD3.5m.


The following five models are evaluated. Note that these models
include both convolutional and transformer-based architectures.
\begin{enumerate}
    \item \textbf{ResNet-18}~\cite{he2016resnet} is a lightweight 18-layer
          residual network, providing a low-capacity CNN baseline.
          Feature dimension: 512; pretrained on ImageNet-1K.
    \item \textbf{ResNet-50}~\cite{he2016resnet} is the 50-layer variant residual network,
          providing a medium-capacity CNN baseline.
          Feature dimension: 2,048; pretrained on ImageNet-1K.
    \item \textbf{EfficientNet-B0}~\cite{tan2019efficientnet} is a
          compound-scaled CNN, optimized for efficiency.
          Feature dimension: 1,280; pretrained on ImageNet-1K.
    \item \textbf{ConvNeXt-Base}~\cite{liu2022convnext} is a modern CNN
          architecture incorporating design principles from Vision Transformers,
          serving as a bridge between pure CNN and transformer families.
          Feature dimension: 1,024; pretrained on ImageNet-1K.
    \item \textbf{CLIP ViT-L/14}~\cite{radford2021clip} is a large
          Vision Transformer trained on~400M image-text pairs with a
          contrastive language-image objective, and a feature dimension of~768.
          This backbone is particularly relevant because its representations
          are sensitive to semantic and stylistic content rather than
          low-level texture statistics.
\end{enumerate}

\subsubsection{Dataset Splits, Pruning, and Preprocessing}\label{sec:splits}
 
The AI-ArtBench dataset~\cite{silva_artbrain_2024} is built on top of
ArtBench-10~\cite{liao_artbench_2022}, which contains~6,000 images per art
style (60,000 human artwork samples in total). AI-ArtBench splits the human
portion into a train split of~5,000 images per style (50,000 total) and
a test split of~1,000 images per style (10,000 total), adding AI-generated
images from LDM and SD2.1 to each split. Table~\ref{tab:aiartbench_original}
reproduces the full dataset composition from the original
paper~\cite{silva_artbrain_2024}. The~30,000-image test split is used
directly as the in-distribution (ID) test set in this study and is not
used during training or validation.
 
\begin{table}[!htb]
\centering
\caption{Original AI-ArtBench dataset composition}
\label{tab:aiartbench_original}
\begin{adjustbox}{scale=0.85}
\begin{tabular}{lccc}
\toprule
\textbf{Source} & \textbf{Styles} & \textbf{Train} & \textbf{Test} \\
\midrule
Latent Diffusion (LDM) & 10 & \zz52{,}092 & 10,000 \\
Stable Diffusion 2.1 (SD2.1) & 10 & \zz52,923 & 10,000 \\
Human (WikiArt / ArtBench-10) & 10 & \zz50,000 & 10,000 \\
\midrule
\textbf{Total} & \textbf{30} & \textbf{155,015} & \textbf{30,000} \\
\midrule
\end{tabular}
\end{adjustbox}
\end{table}
 
The prompt-aligned SD3.5m OOD evaluation requires a held-out set of human
artwork samples that share no overlap with training data. These images serve as
both the source for reverse prompting (Section~\ref{sec:clip_interrogator})
and the human-class reference in the final OOD test set. To obtain this set,
1,000 human images per style (10,000 in total) were extracted from 
the~50,000-image human training pool. The remaining human training pool contains 
exactly~4,000 images per style (40,000 total).
 
After OOD extraction, the training pool consists of~40,000 human images,
52,092 LDM images, and~52,923 SD2.1 images. The AI generator counts are
substantially larger than the human count and vary unevenly across styles:
LDM per-style counts range from~4,844 to~5,504 and SD2.1 per-style counts
range from~5,052 to~5,455. Table~\ref{tab:stylewise_pruning} shows the
full per-style breakdown before and after pruning. Without pruning, the
combined training set would contain~105,015 AI images against~40,000 human images, a~2.6:1 binary-class imbalance that would bias detectors toward the majority AI class and conflate style-specific detection difficulty with class frequency effects.
 
To remove the source-level and style-level frequency differences, exactly~4,000 images per style were retained for each AI
source class by random sampling with a fixed seed.
This removed~12,092 excess LDM images and~12,923 excess SD2.1 images,
totaling~25,015 pruned images. The resulting corpus contains
$3 \times 4{,}000 \times 10 = 120{,}000$ images with equal
representation across all style-source combinations.
 
\begin{table}[!htb]
\centering
\caption{Per-style image counts in the training pool before and after pruning}
\label{tab:stylewise_pruning}
\begin{adjustbox}{scale=0.735}
\begin{tabular}{l ccc c ccc c}
\midrule
\multirow{2}{*}{\raisebox{-3.0pt}{\textbf{Style}}} & \multicolumn{4}{c}{\textbf{Before pruning}} &
   \multicolumn{4}{c}{\textbf{After pruning}} \\
\cmidrule(lr){2-5}\cmidrule(lr){6-9}
  &
  \textbf{Human} & \textbf{LDM} & \textbf{SD2.1} & \textbf{Total} &
  \textbf{Human} & \textbf{LDM} & \textbf{SD2.1} & \textbf{Total} \\
\midrule
Art Nouveau        & 4{,}000 & 4{,}992 & 5{,}384 & 14{,}376 &
                     4{,}000 & 4{,}000 & 4{,}000 & 12{,}000 \\
Baroque            & 4{,}000 & 4{,}960 & 5{,}052 & 14{,}012 &
                     4{,}000 & 4{,}000 & 4{,}000 & 12{,}000 \\
Expressionism      & 4{,}000 & 5{,}116 & 5{,}304 & 14{,}420 &
                     4{,}000 & 4{,}000 & 4{,}000 & 12{,}000 \\
Impressionism      & 4{,}000 & 5{,}208 & 5{,}388 & 14{,}596 &
                     4{,}000 & 4{,}000 & 4{,}000 & 12{,}000 \\
Post-Impressionism & 4{,}000 & 4{,}844 & 5{,}360 & 14{,}204 &
                     4{,}000 & 4{,}000 & 4{,}000 & 12{,}000 \\
Realism            & 4{,}000 & 5{,}232 & 5{,}248 & 14{,}480 &
                     4{,}000 & 4{,}000 & 4{,}000 & 12{,}000 \\
Renaissance        & 4{,}000 & 5{,}416 & 5{,}060 & 14{,}476 &
                     4{,}000 & 4{,}000 & 4{,}000 & 12{,}000 \\
Romanticism        & 4{,}000 & 5{,}436 & 5{,}455 & 14{,}891 &
                     4{,}000 & 4{,}000 & 4{,}000 & 12{,}000 \\
Surrealism         & 4{,}000 & 5{,}384 & 5{,}364 & 14{,}748 &
                     4{,}000 & 4{,}000 & 4{,}000 & 12{,}000 \\
Ukiyo-e            & 4{,}000 & 5{,}504 & 5{,}308 & 14{,}812 &
                     4{,}000 & 4{,}000 & 4{,}000 & 12{,}000 \\
\midrule
\textbf{Total}  & \textbf{40{,}000} & \textbf{52{,}092} & \textbf{52{,}923} &
                  \textbf{145{,}015} &
                  \textbf{40{,}000} & \textbf{40{,}000} & \textbf{40{,}000} &
                  \textbf{120{,}000} \\
\bottomrule
\end{tabular}
\end{adjustbox}
\end{table}
 
The before-pruning column in Table~\ref{tab:stylewise_pruning} makes the
imbalance concrete: the combined pre-pruning pool contains~105,015 AI
images versus~40,000 human images. Within the AI classes, per-style
variation is as high as~660 images for LDM (4,844 in Post-Impressionism
versus~5,504 in Ukiyo-e) and~403 images for SD2.1 (5,052 in Baroque versus~5,455 in Romanticism). 
Pruning equalizes the three source categories and removes within-source style skew, ensuring that no style-source combination contributes disproportionately to detector training.
 
The~120,000-image balanced corpus was divided into a training split
of~105,000 images (3,500 per style per source class) and a validation
split of~15,000 images (500 per style per source class), maintaining the source balance within each split. Images were assigned to each split using a fixed random seed. The validation split is used
exclusively for epoch selection and threshold optimization and is not
used for final evaluation. The complete dataset composition across all
splits is summarized in Tables~\ref{tab:id_splits} and~\ref{tab:ood_splits}. 
Overall, the final in-distribution dataset was split into~70\%\ training, 10\%\ validation, and~20\%\ ID test.
 

\begin{table}[!htb]
\centering
\caption{In-distribution (ID) dataset composition across all splits}
\label{tab:id_splits}
\begin{adjustbox}{scale=0.85}
\begin{tabular}{lrrrr}
\toprule
\textbf{Split} & \textbf{Human} & \textbf{LDM} &
  \textbf{SD2.1} & \textbf{Total} \\
\midrule
Training   & 35{,}000 & 35{,}000 & 35{,}000 & 105{,}000 \\
Validation &  5{,}000 &  5{,}000 &  5{,}000 &  15{,}000 \\
ID Test    & 10{,}000 & 10{,}000 & 10{,}000 &  30{,}000 \\
\midrule
\textbf{Total} &
  \textbf{50{,}000} & \textbf{50{,}000} &
  \textbf{50{,}000} & \textbf{150{,}000} \\
\bottomrule
\end{tabular}
\end{adjustbox}
\end{table}

\begin{table}[!htb]
\centering
\caption{Out-of-distribution (OOD) evaluation set composition}
\label{tab:ood_splits}
\begin{adjustbox}{scale=0.85}
\begin{tabular}{lr}
\toprule
\textbf{Source} & \textbf{Images} \\
\midrule
Human (held-out)  & 10{,}000 \\
AI (SD3.5m, generated)     & 10{,}000 \\
\midrule
\textbf{Total}             & \textbf{20{,}000} \\
\bottomrule
\end{tabular}
\end{adjustbox}
\end{table}
 
All preprocessing performed is deterministic and no data augmentation is applied at
any stage. Each image is loaded using the Pillow Python imaging library (PIL), converted to RGB, and passed through the same deterministic backbone-specific preprocessing pipeline before feature extraction. For the four CNN backbones 
(ResNet-18, ResNet-50, EfficientNet-B0 and ConvNeXt-Base), images are resized 
to~$224\times 224$ pixels using bicubic interpolation, converted to tensors,
and normalized using the ImageNet training set mean
($\mu = [0.485, 0.456, 0.406]$) and standard deviation
($\sigma = [0.229, 0.224, 0.225]$)~\cite{russakovsky2015imagenet}. For
CLIP ViT-L/14, the HuggingFace processor associated with
$$
    \texttt{openai/clip-vit-large-patch14} 
$$
is used to prepare images for
feature extraction, applying the checkpoint’s default resize/crop and
normalization pipeline. The same backbone-specific deterministic
preprocessing is used for all splits including training, validation, ID
test and OOD test.

\subsubsection{Feature Extraction}\label{sec:feature_extraction}

Each backbone is used as a frozen feature extractor: the final
classification layer is removed so that a forward pass returns a
fixed-length feature vector for each image. For CLIP ViT-L/14, the
projected image embedding is used, consistent with the space in which
CLIP was pretrained. Table~\ref{tab:backbone_features} lists the
pretrained weights and output feature dimension for each backbone. 
CNN backbones are loaded using torchvision and CLIP ViT-L/14
is loaded using HuggingFace Transformers.

\begin{table}[!htb]
\centering
\caption{Pretrained weights and output feature dimension for each backbone}\label{tab:backbone_features}
\begin{adjustbox}{scale=0.85}
\begin{tabular}{lcc}
\toprule
\multirow{2}{*}{\textbf{Model}} & \textbf{Pretrained} & \textbf{Feature} \\
  & \textbf{weights} & \textbf{dimension} \\
\midrule
ResNet-18       & ImageNet-1K (V1) & \zzc512   \\
ResNet-50       & ImageNet-1K (V2) & 2{,}048 \\
EfficientNet-B0 & ImageNet-1K (V1) & 1{,}280 \\
ConvNeXt-Base   & ImageNet-1K (V1) & 1{,}024 \\
CLIP ViT-L/14   & \texttt{openai/clip-vit-large-patch14} & \zzc768 \\
\bottomrule
\end{tabular}
\end{adjustbox}
\end{table}

Features are extracted once per backbone per split and saved to disk.
All subsequent head training reads directly from these cached feature
files, reducing the per-epoch training cost to a single linear layer
pass and allowing rapid iteration across training epochs.
 
\subsubsection{Classification Head Training}\label{sec:head_training}

A single linear layer maps the frozen backbone feature vector~$\mathbf{x} \in \mathbb{R}^{D}$ 
to a scalar logit according to
\begin{equation}\label{eq:logit}
    z = \mathbf{w}^{\top}\mathbf{x} + b, \mbox{\ where\ }
    \mathbf{w} \in \mathbb{R}^{D} \mbox{\ and\ } b \in \mathbb{R}
\end{equation}
At inference, the logit is converted to an AI probability score
via the sigmoid function
$$
    p_{\AI} = \sigma(z) = \frac{1}{1 + e^{-z}} \in (0, 1)
$$
The head contains no hidden layers and no dropout. This linear
probe design is intentional because the backbone is frozen, the trainable
component is restricted to a lightweight classifier operating on a
high-dimensional pretrained feature space. This keeps the training setup
simple, reduces the risk of overfitting in the head, and makes comparisons
across backbone families easier to interpret, since most of the
discriminative capacity comes from the pretrained representation rather
than from a more expressive task-specific head.

The head is optimized using weighted binary cross-entropy with logits.
For a mini-batch of $N$ samples with logits~$\{z_i\}$ and binary labels~$\{y_i\} \in \{0, 1\}$, the loss is
\begin{equation}\label{eq:bce}
    \mathcal{L} = -\frac{1}{N} \sum_{i=1}^{N} w_{y_i}
    \Bigl[ y_i \log \sigma(z_i) + (1 - y_i) \log (1 - \sigma(z_i)) \Bigr]
\end{equation}
where~$w_0$ and~$w_1$ are per-class weights computed from the
training split as
$$
    w_c = \frac{N}{2 \cdot N_c}, \mbox{ for } c \in \{0, 1\}
$$
where~$N_c$ denotes the number of training samples in class~$c$.
This inverse-frequency weighting increases the contribution of the minority
class and reduces bias toward the majority class. In the completed binary
training split, the class counts are~35,000 human and~70,000 AI samples,
which correspond to weights of~1.5 for the human class and~0.75 for the AI
class. The loss is computed directly from logits using
\texttt{F.binary\_cross\_entropy\_with\_logits}, which combines the
sigmoid and cross-entropy terms in a single numerically stable operation.

AdamW~\cite{loshchilov2019adamw} is used with a fixed learning rate 
of~$1 \times 10^{-3}$ and weight decay of~$1 \times 10^{-4}$.
AdamW is preferred over standard Adam because it decouples weight decay
from the gradient update, yielding more consistent regularization in the
presence of adaptive learning rates. In this setting, weight decay acts as
L2-style regularization on the head weights and helps limit overfitting in
the linear classifier.

Model training runs for a maximum of ten epochs. At the end of each epoch, the
head is evaluated on the validation split using a fixed threshold of~0.5. 
The checkpoint with the highest validation balanced accuracy is retained.
Balanced accuracy is computed as
$$
\text{Balanced Accuracy} = \frac{1}{2}
\left(\frac{\text{TP}}{\text{TP}+\text{FN}} +
      \frac{\text{TN}}{\text{TN}+\text{FP}}\right)
$$
and is used as the model-selection criterion rather than
standard accuracy because it gives equal weight to the true positive rate and true
negative rate.
This is more informative than raw accuracy under class imbalance
and better reflects the need to control both missed AI images and
misclassified human artwork samples. The training configuration also includes
patience-based early stopping with a patience of three epochs, although in
runs that continued to improve, training proceeded to the full ten-epoch
budget. After the best checkpoint is selected, a separate threshold sweep is
performed on the validation predictions.
These configuration parameters are summarized in Table~\ref{tab:training_config}.

\begin{table}[!htb]
\centering
\caption{Shared linear head training configuration}
\label{tab:training_config}
\begin{adjustbox}{scale=0.85}
\begin{tabular}{ll}
\toprule
\textbf{Parameter}         & \textbf{Value} \\
\midrule
Maximum epochs             & 10 \\
Loss function              & Weighted binary cross-entropy with logits: equation~\eref{eq:bce} \\
Optimizer                  & AdamW \\
Learning rate              & $1 \times 10^{-3}$ \\
Weight decay               & $1 \times 10^{-4}$ \\
Epoch selection criterion  & Best validation balanced accuracy \\
Image resolution           & 224 $\times$ 224 \\
Interpolation              & Bicubic \\
Normalization              & ImageNet mean and standard deviation \\
Backbone weights           & Frozen (no fine-tuning) \\
Classification head        & Single linear layer, logit output: equation~\eref{eq:logit} \\
Task                       & Binary: human ($y=0$) versus AI ($y=1$) \\
\bottomrule
\end{tabular}
\end{adjustbox}
\end{table}
 
\subsubsection{Threshold Selection}\label{sec:threshold}

The sigmoid output~$p_{\AI} \in [0, 1]$ must be mapped to a binary prediction using a threshold. While the standard choice of a threshold is~0.5, such a threshold would be suboptimal when the score distribution is not centered
around~0.5 or when the relative impact of false positives and false
negatives differs. To select an operating threshold for each model, a sweep over~99
linearly spaced candidate values from~0.01 to~0.99 was performed on the
validation set. For each candidate threshold, binary predictions were derived and
validation balanced accuracy was computed, with F1 score used as a
tie-breaker. The threshold that maximized validation balanced accuracy was
selected and locked for all subsequent evaluations on the ID test and OOD
sets. This threshold is not adjusted after selection, preventing
threshold overfitting and ensuring that the reported test and OOD metrics
reflect realistic deployment conditions.

Selecting thresholds using validation balanced accuracy provides a consistent basis for comparing the models under the assumption that AI-image recall and human-image specificity are equally important. However, the most appropriate operating threshold in practice may depend on the defensive context and the relative costs of false positives and false negatives. For example, a news-verification or high-risk fraud-screening workflow may prioritize recall and tolerate a higher false-positive rate so that fewer synthetic images escape review. In contrast, an art-attribution or authenticity-assessment system may require a very low false-positive rate to reduce the risk of incorrectly labeling human-created artwork as synthetic. The thresholds used in this study are therefore intended for controlled cross-model and cross-generator comparison rather than as universally optimal deployment thresholds.
 
\subsubsection{Evaluation Protocol}\label{sec:eval_protocol}

Detectors were evaluated on three disjoint sets: the validation set (used
for epoch selection and threshold selection), the ID test set (drawn from
the same generator distribution as training), and the OOD evaluation set (unseen SD3.5m dataset). For each model on each split, the following metrics were computed at the validation-selected threshold, with AI treated as the positive class: balanced accuracy, precision, recall, F1 score, Matthews
correlation coefficient (MCC)~\cite{baldi2000assessing, matthews1975comparison}, false positive rate (FPR), false negative
rate (FNR), ROC-AUC, and PR-AUC. Here, recall corresponds to the detector’s coverage of AI-generated images, FNR corresponds to the AI-image miss rate or evasion-like failure rate, and FPR corresponds to the false-alarm rate on human artwork. Style-wise breakdowns were also computed
to identify which artistic styles contributed most strongly to model
failures. In addition, source-wise analyses were performed for the in-distribution setting.

The main result of interest is detector performance on the OOD set, especially the false negative rate on SD3.5m images. An OOD AI image is
classified as AI if~$p_{\AI} \geq \tau_{\text{val}}$, 
where~$\tau_{\text{val}}$ is the threshold selected on the validation set. The
drop in balanced accuracy between the ID test set and OOD set is used
to calculate how much each detector degrades under generator shift from the
training generators (LDM and SD2.1) to SD3.5m. All experiments were run on
the hardware described in Section~\ref{sec:hw_sw}, and the pipeline was
designed to support reproducible evaluation through fixed random seeds,
frozen backbone weights, deterministic preprocessing, and cached feature
bundles.

\section{Experimental Results and Analysis}\label{chap:results}

This section reports the experimental outcomes for AI-art detection and provides analysis of the results. Section~\ref{sec:training_dynamics} describes training dynamics
and the selected classification thresholds. Section~\ref{sec:id_performance}
reports in-distribution performance on the validation and ID test splits. Then in
Section~\ref{sec:ood_results}, we present cross-generator OOD evaluation on the prompt-aligned SD3.5m test set, including
the ID-to-OOD performance gap. Section~\ref{sec:stylewise_ood} provides a
per-style breakdown of OOD performance, Section~\ref{sec:sourcewise_id} reports
source-wise accuracy on the ID test set, and Section~\ref{sec:analysis} analyzes
the results using Grad-CAM spatial attribution maps.

Throughout our experiments, detector models are identified by their backbone name:
ResNet-18, ResNet-50, EfficientNet-B0, ConvNeXt-Base, and CLIP ViT-L/14.
The primary evaluation metric is balanced accuracy, which weights the true
positive rate and true negative rate equally. When a single ranking is needed, 
balanced accuracy is used as the primary criterion, with F1 score,
recall, and ROC-AUC provided for various experiments as secondary criteria.

\subsection{Model Training and Threshold Selection}\label{sec:training_dynamics}

The detectors were trained for ten epochs using the frozen-backbone,
linear-head setup described in Section~\ref{sec:head_training}.
Table~\ref{tab:training_dynamics} reports the validation F1 score and
validation balanced accuracy at the end of each epoch for every model.
The best-epoch checkpoint for each model was selected based on the highest
validation balanced accuracy achieved across all ten epochs.

\begin{table}[!htb]
\centering
\caption{Validation F1 score and balanced accuracy by epoch}
\label{tab:training_dynamics}
\begin{adjustbox}{scale=0.7125}
\begin{tabular}{c cc cc cc cc cc}
\toprule
\multirow{2}{*}{\raisebox{-3.0pt}{\textbf{Epoch}}} & \multicolumn{2}{c}{\textbf{ResNet-18}} &
   \multicolumn{2}{c}{\textbf{ResNet-50}} &
   \multicolumn{2}{c}{\textbf{EffNet-B0}} &
   \multicolumn{2}{c}{\textbf{ConvNeXt-Base}} &
   \multicolumn{2}{c}{\textbf{CLIP ViT-L/14}} \\ 
   	\cmidrule(lr){2-3}\cmidrule(lr){4-5}\cmidrule(lr){6-7}\cmidrule(lr){8-9}\cmidrule(lr){10-11}
 & \textbf{F1} & \textbf{BalAcc} & \textbf{F1} & \textbf{BalAcc} & \textbf{F1} & \textbf{BalAcc} 
 & \textbf{F1} & \textbf{BalAcc} & \textbf{F1} & \textbf{BalAcc} \\
\midrule
 \zz1 & 0.871 & 0.835 & 0.924 & 0.897 & 0.927 & 0.903 & 0.929 & 0.904 & 0.979 & 0.973 \\
 \zz2 & 0.907 & 0.874 & 0.948 & 0.927 & 0.945 & 0.923 & 0.956 & 0.935 & 0.991 & 0.988 \\
 \zz3 & 0.922 & 0.892 & 0.957 & 0.940 & 0.951 & 0.932 & 0.964 & 0.948 & 0.994 & 0.992 \\
 \zz4 & 0.927 & 0.902 & 0.962 & 0.948 & 0.956 & 0.938 & 0.968 & 0.955 & 0.996 & 0.995 \\
 \zz5 & 0.931 & 0.908 & 0.966 & 0.954 & 0.958 & 0.943 & 0.972 & 0.960 & 0.997 & 0.996 \\
 \zz6 & 0.936 & 0.913 & 0.970 & 0.958 & 0.961 & 0.946 & 0.975 & 0.963 & 0.997 & 0.997 \\
 \zz7 & 0.939 & 0.917 & 0.972 & 0.961 & 0.961 & 0.948 & 0.977 & 0.967 & 0.998 & 0.997 \\
 \zz8 & 0.940 & 0.920 & 0.973 & 0.964 & 0.962 & 0.950 & 0.978 & 0.969 & 0.998 & 0.997 \\
 \zz9 & 0.942 & 0.922 & 0.974 & 0.965 & 0.964 & 0.952 & 0.980 & 0.972 & 0.998 & 0.998 \\
10 & \textbf{0.945} & \textbf{0.924} &
    \textbf{0.976} & \textbf{0.966} &
    \textbf{0.966} & \textbf{0.953} &
    \textbf{0.981} & \textbf{0.973} &
    \textbf{0.998} & \textbf{0.998} \\
\bottomrule
\end{tabular}
\end{adjustbox}
\end{table}

\begin{figure}[!htb]
\centering
\includegraphics[scale=0.45]{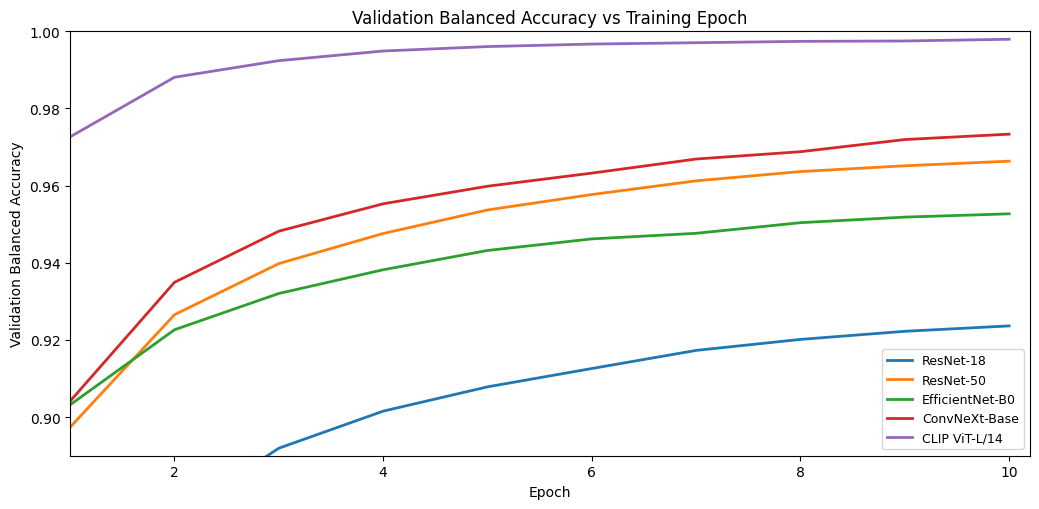}
\caption{Validation balanced accuracy as a function of training epoch}
\label{fig:training_curves}
\end{figure}

CLIP ViT-L/14 converges faster than all CNN-based models by
epoch~1 it already achieves a validation F1 of~0.979 and a balanced accuracy
of~0.973, whereas ResNet-18 does not reach these values by epoch~10. This suggests that
the semantic representations learned by CLIP through contrastive language-image
pretraining transfer more efficiently to the binary AI-art detection task
than convolutional features pretrained on image classification alone. Among the
CNN backbones, ConvNeXt-Base converges faster and to a higher plateau than the
other three, reflecting its larger capacity and modern architectural design.
ResNet-18 shows the slowest convergence among the detectors. 
Figure~\ref{fig:training_curves} shows the validation balanced accuracy
trajectory for the detector models across ten epochs.

Following training, the optimal decision threshold was selected for each model
as described in Section~\ref{sec:threshold}. Table~\ref{tab:thresholds} reports the selected threshold for
each model. These values range from~0.50 for CLIP ViT-L/14 to~0.58 for
EfficientNet-B0. The CLIP threshold of~0.50 indicates the default midpoint is
already optimal. In contrast, EfficientNet-B0's higher threshold of~0.58 suggests
its raw probabilities are slightly compressed near the center, requiring a shift
toward the AI side to maximize the balanced accuracy.

\begin{table}[!htb]
\centering
\caption{Decision threshold selected and corresponding validation and ID test balanced
accuracy at the threshold}\label{tab:thresholds}
\begin{adjustbox}{scale=0.85}
\begin{tabular}{lccc}
\toprule
\multirow{2}{*}{\raisebox{-3.0pt}{\textbf{Model}}} & 
	\multirow{2}{*}{\raisebox{-3.0pt}{\textbf{Threshold}}} & 
	\multicolumn{2}{c}{\textbf{Balanced accuracy}} \\ \cmidrule{3-4}
& &  \textbf{Validation} & \textbf{ID Test} \\ 
\midrule
CLIP ViT-L/14   & 0.50 & 0.9980 & 0.9969 \\
ConvNeXt-Base   & 0.53 & 0.9745 & 0.9727 \\
ResNet-50       & 0.55 & 0.9672 & 0.9637 \\
EfficientNet-B0 & 0.58 & 0.9538 & 0.9524 \\
ResNet-18       & 0.53 & 0.9246 & 0.9260 \\
\bottomrule
\end{tabular}
\end{adjustbox}
\end{table}

\begin{figure}[!htb]
\centering
\includegraphics[scale=0.5]{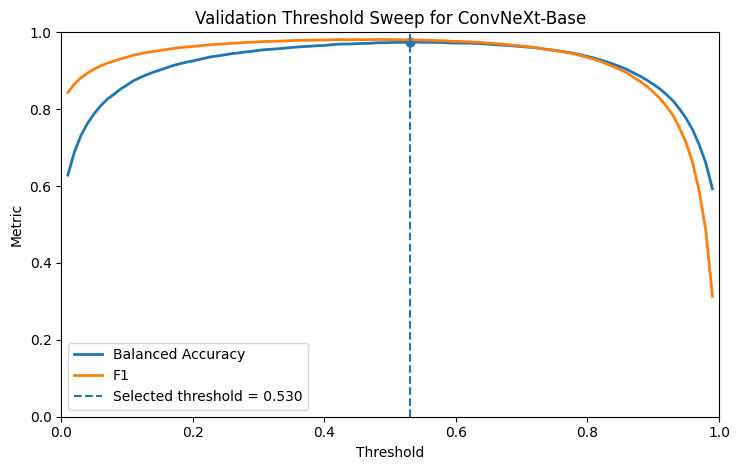}
\caption{Validation threshold sweep for ConvNeXt-Base}
\label{fig:threshold_sweep}
\end{figure}

Figure~\ref{fig:threshold_sweep} illustrates the threshold sweep for
ConvNeXt-Base, showing how validation balanced accuracy varies across the~99
candidate threshold values and confirming that the selected value of~0.53 is
indeed at the peak.

\subsection{In-Distribution Performance}\label{sec:id_performance}

Table~\ref{tab:id_results} reports evaluation results for the detectors on the validation and ID test splits, using the thresholds selected from the validation set. In-distribution performance measures how well each detector identifies images from the same generator families used during training.


\begin{table}[!htb]
\centering
\caption{In-distribution performance on the validation split (Val)
and ID test split (IDT); here $\mbox{Bal}=\mbox{balanced accuracy}$, $\mbox{Prec}=\mbox{precision}$,
$\mbox{Rec}=\mbox{recall}$, $\mbox{F1}=\mbox{F1 score}$, $\mbox{FNR}=\mbox{false negative rate}$,
$\mbox{FPR}=\mbox{false positive rate}$, $\mbox{MCC}=\mbox{Matthews correlation coefficient}$,
$\mbox{AUC}=\mbox{ROC-AUC}$, $\mbox{PR}=\mbox{PR-AUC}$, and all metrics computed at the
validation-selected threshold as per Table~\ref{tab:thresholds}}
\label{tab:id_results}
\begin{adjustbox}{scale=0.735}
\begin{tabular}{llrrrrrrrrr}
\toprule
\textbf{Model} & \textbf{Split} &
  \multicolumn{1}{c}{\textbf{Bal}} & \multicolumn{1}{c}{\textbf{Prec}} & \multicolumn{1}{c}{\textbf{Rec}} &
  \multicolumn{1}{c}{\textbf{F1}} & \multicolumn{1}{c}{\textbf{FNR}} & \multicolumn{1}{c}{\textbf{FPR}} &
  \multicolumn{1}{c}{\textbf{MCC}} & \multicolumn{1}{c}{\textbf{AUC}} & \multicolumn{1}{c}{\textbf{PR}} \\
\midrule
\multirow{2}{*}{CLIP ViT-L/14}
  & Val & 0.9980 & 0.9992 & 0.9975 & 0.9983 & 0.0025 & 0.0016 & 0.9951 & 1.0000 & 1.0000 \\
  & IDT & 0.9969 & 0.9984 & 0.9971 & 0.9977 & 0.0030 & 0.0032 & 0.9932 & 0.9999 & 1.0000 \\
\midrule
\multirow{2}{*}{ConvNeXt-Base}
  & Val & 0.9745 & 0.9870 & 0.9745 & 0.9807 & 0.0255 & 0.0256 & 0.9431 & 0.9962 & 0.9980 \\
  & IDT & 0.9727 & 0.9863 & 0.9726 & 0.9794 & 0.0275 & 0.0271 & 0.9391 & 0.9962 & 0.9980 \\
\midrule
\multirow{2}{*}{ResNet-50}
  & Val & 0.9672 & 0.9847 & 0.9645 & 0.9745 & 0.0355 & 0.0300 & 0.9254 & 0.9940 & 0.9968 \\
  & IDT & 0.9637 & 0.9821 & 0.9624 & 0.9721 & 0.0377 & 0.0350 & 0.9185 & 0.9939 & 0.9968 \\
\midrule
\multirow{2}{*}{EfficientNet-B0}
  & Val & 0.9538 & 0.9795 & 0.9472 & 0.9631 & 0.0528 & 0.0396 & 0.8939 & 0.9901 & 0.9946 \\
  & IDT & 0.9524 & 0.9776 & 0.9482 & 0.9627 & 0.0518 & 0.0435 & 0.8922 & 0.9900 & 0.9949 \\
\midrule
\multirow{2}{*}{ResNet-18}
  & Val & 0.9246 & 0.9595 & 0.9274 & 0.9432 & 0.0726 & 0.0782 & 0.8362 & 0.9767 & 0.9868 \\
  & IDT & 0.9260 & 0.9603 & 0.9289 & 0.9443 & 0.0712 & 0.0768 & 0.8393 & 0.9777 & 0.9874 \\
\bottomrule
\end{tabular}
\end{adjustbox}
\end{table}

All models perform well on the in-distribution test set, the ranking generally follows the strength and capacity of the various models. CLIP ViT-L/14 is the best in-distribution detector, as it achieves an ID test balanced accuracy of~0.9969, an F1 score of~0.9977, and a ROC-AUC of~0.9999. Its false positive rate and false negative rate are both~0.003, which implies that CLIP ViT-L/14 makes fewer than~100 errors across the~30,000 ID test images. ConvNeXt-Base is the second best model with an ID test balanced accuracy of~0.9727. The other CNN models perform close to each other: ResNet-50 reaches~0.9637, EfficientNet-B0 achieves~0.9524, and ResNet-18 is at~0.9260. The validation and ID test results are also very consistent for all models,
indicating that the threshold selection did not overfit to the validation split.

\begin{figure}[!htb]
\centering
\begin{tabular}{cc}
    \adjustbox{scale=0.8}{%
\begin{tikzpicture}[scale=1.0]
    \begin{axis}[
        width=4.5cm,
        height=4.5cm,
	colormap={bluewhite}{color=(white) rgb255=(100,149,237)},
        xticklabels={
        \texttt{Human},
        \texttt{AI},
        },
        xtick={0,...,1},
        xtick style={draw=none},
	xticklabel style={font=\tt,scale=1.0},
        yticklabels={
        \texttt{Human},
        \texttt{AI},
        },
        ytick={0,...,1},
        ytick style={draw=none},
        enlargelimits=false,
        yticklabel style={xshift=0pt,font=\tt,scale=1.0},
        colorbar,
        colorbar/width=2.5mm,
        colorbar style={
            ytick={0,5000,10000,15000,20000},
            yticklabels={0,5000,10000,15000,20000},
            ytick style={draw=none},
	    scaled y ticks=false,
            yticklabel={\pgfmathprintnumber\tick},
            yticklabel style={font=\tt, scale=0.9,
            		/pgf/number format/fixed,
			/pgf/number format/fixed zerofill,
			/pgf/number format/precision=0}
        },
        point meta min=0,
        point meta max=20000,
        nodes near coords={\pgfmathprintnumber\pgfplotspointmeta},
        nodes near coords black white/.style={
            small value/.style={
                yshift=-7pt,
                text=black,
                /pgf/number format/fixed,
                /pgf/number format/precision=0,
                /pgf/number format/zerofill=true,
                scale=1.0,
            },
            large value/.style={
                yshift=-7pt,
                text=white,
                /pgf/number format/fixed,
                /pgf/number format/precision=0,
                /pgf/number format/zerofill=true,
                scale=1.0,
            },
            every node near coord/.style={
                check for zero/.code={
                    \pgfmathfloatifflags{\pgfplotspointmeta}{0}{
                        \pgfkeys{/tikz/coordinate}
                    }{
                        \begingroup
                        \pgfkeys{/pgf/fpu}
                        \pgfmathparse{\pgfplotspointmeta<#1}
                        \global\let\result=\pgfmathresult
                        \endgroup
                        %
                        %
                        \pgfmathfloatcreate{1}{1.0}{0}
                        \let\ONE=\pgfmathresult
                        \ifx\result\ONE
                            \pgfkeysalso{/pgfplots/small value}
                        \else
                            \pgfkeysalso{/pgfplots/large value}
                        \fi
                    }
                },
                check for zero,
            },
        },
        nodes near coords black white=10000,
    ]
        \addplot[
            matrix plot,
            mesh/cols=2,
            point meta=explicit,draw=gray
        ] table [meta=C] {
            x y C
0 0 9968
1 0 32
0 1 59
1 1 19941
         };
    \end{axis}
\end{tikzpicture}
    }
    &
    \adjustbox{scale=0.8}{%
\begin{tikzpicture}[scale=1.0]
    \begin{axis}[
        width=4.5cm,
        height=4.5cm,
	colormap={bluewhite}{color=(white) rgb255=(100,149,237)},
        xticklabels={
        \texttt{Human},
        \texttt{AI},
        },
        xtick={0,...,1},
        xtick style={draw=none},
	xticklabel style={font=\tt,scale=1.0},
        yticklabels={
        \texttt{Human},
        \texttt{AI},
        },
        ytick={0,...,1},
        ytick style={draw=none},
        enlargelimits=false,
        yticklabel style={xshift=0pt,font=\tt,scale=1.0},
        colorbar,
        colorbar/width=2.5mm,
        colorbar style={
            ytick={0,5000,10000,15000,20000},
            yticklabels={0,5000,10000,15000,20000},
            ytick style={draw=none},
	    scaled y ticks=false,
            yticklabel={\pgfmathprintnumber\tick},
            yticklabel style={font=\tt, scale=0.9,
            		/pgf/number format/fixed,
			/pgf/number format/fixed zerofill,
			/pgf/number format/precision=0}
        },
        point meta min=0,
        point meta max=20000,
        nodes near coords={\pgfmathprintnumber\pgfplotspointmeta},
        nodes near coords black white/.style={
            small value/.style={
                yshift=-7pt,
                text=black,
                /pgf/number format/fixed,
                /pgf/number format/precision=0,
                /pgf/number format/zerofill=true,
                scale=1.0,
            },
            large value/.style={
                yshift=-7pt,
                text=white,
                /pgf/number format/fixed,
                /pgf/number format/precision=0,
                /pgf/number format/zerofill=true,
                scale=1.0,
            },
            every node near coord/.style={
                check for zero/.code={
                    \pgfmathfloatifflags{\pgfplotspointmeta}{0}{
                        \pgfkeys{/tikz/coordinate}
                    }{
                        \begingroup
                        \pgfkeys{/pgf/fpu}
                        \pgfmathparse{\pgfplotspointmeta<#1}
                        \global\let\result=\pgfmathresult
                        \endgroup
                        %
                        %
                        \pgfmathfloatcreate{1}{1.0}{0}
                        \let\ONE=\pgfmathresult
                        \ifx\result\ONE
                            \pgfkeysalso{/pgfplots/small value}
                        \else
                            \pgfkeysalso{/pgfplots/large value}
                        \fi
                    }
                },
                check for zero,
            },
        },
        nodes near coords black white=10000,
    ]
        \addplot[
            matrix plot,
            mesh/cols=2,
            point meta=explicit,draw=gray
        ] table [meta=C] {
            x y C
0 0 9729
1 0 271
0 1 549
1 1 19451
         };
    \end{axis}
\end{tikzpicture}
    }
    \\
    \adjustbox{scale=0.85}{(a) CLIP ViT-L/14}
    &
    \adjustbox{scale=0.85}{(b) ConvNeXt-Base}
\end{tabular}
\caption{ID test confusion matrices}\label{fig:id_confusion}
\end{figure}
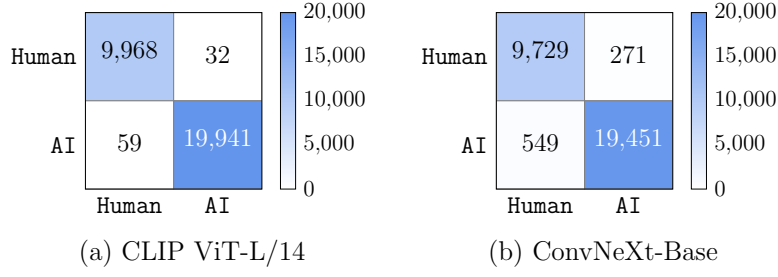

\begin{figure}[!htb]
\centering
\includegraphics[scale=0.5]{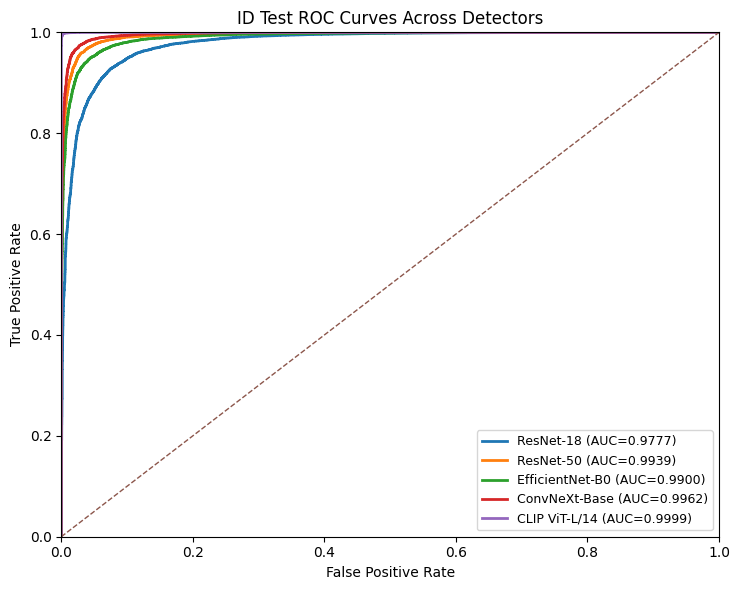}
\caption{ID test ROC curves for all detectors}
\label{fig:id_roc}
\end{figure}

Figure~\ref{fig:id_confusion} shows the ID test confusion matrices for CLIP
ViT-L/14 and ConvNeXt-Base. Figure~\ref{fig:id_roc} overlays the ID test
ROC curves for each of the five models.


\begin{figure}[!htb]
\centering
\begin{tabular}{cc}
    \includegraphics[scale=0.375]{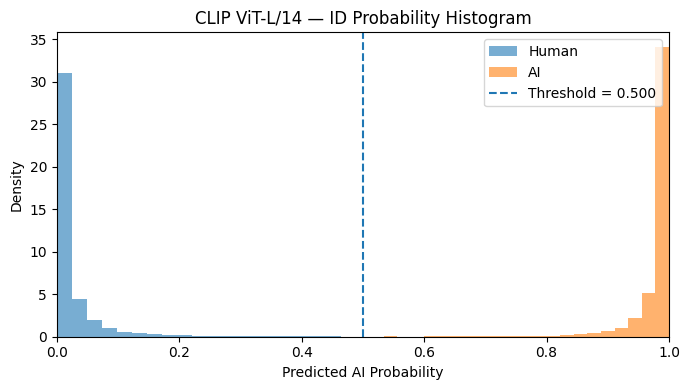}
    &
    \includegraphics[scale=0.375]{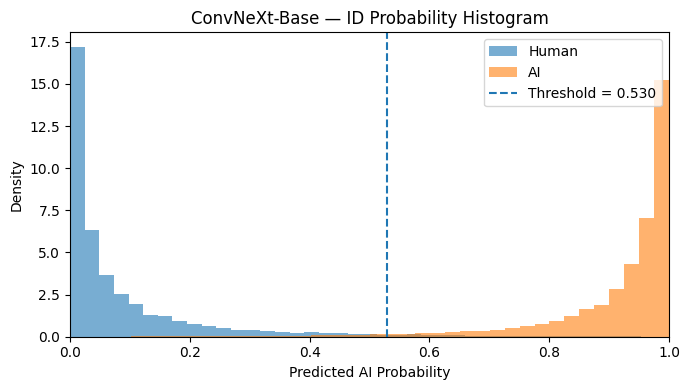}
    \\
    \adjustbox{scale=0.85}{(a) CLIP ViT-L/14}
    &
    \adjustbox{scale=0.85}{(b) ConvNeXt-Base}
\end{tabular}
\caption{ID test predicted-probability histograms}\label{fig:id_prob_hist}
\end{figure}

Figure~\ref{fig:id_prob_hist} shows the ID test probability histograms for
CLIP ViT-L/14 and ConvNeXt-Base. For both models, human images are mostly grouped near~$p_{\AI}=0$, while AI images are mostly grouped near~$p_{\AI}=1$, and there is little overlap around the threshold selected from the validation set. This clear separation indicates that both models can distinguish human and AI images very well on the in-distribution test set.


Table~\ref{tab:id_stylewise_top2} reports per-style ID test balanced accuracy
for CLIP ViT-L/14 and ConvNeXt-Base. All per-style accuracies
are above~0.94 for ConvNeXt-Base, and above~0.994 for CLIP ViT-L/14. 
Surrealism is the most challenging style
for both models. Ukiyo-e is the easiest style for ConvNeXt-Base,
while all styles are near-uniformly easy for CLIP ViT-L/14.


\begin{table}[!htb]
\centering
\caption{Per-style ID test balanced accuracy for CLIP ViT-L/14 and
ConvNeXt-Base}\label{tab:id_stylewise_top2}
\begin{adjustbox}{scale=0.85}
\begin{tabular}{lcc}
\toprule
\textbf{Style} & \textbf{CLIP ViT-L/14} & \textbf{ConvNeXt-Base} \\
\midrule
Art Nouveau        & 0.9978 & 0.9860 \\
Baroque            & 0.9978 & 0.9818 \\
Expressionism      & 0.9955 & 0.9725 \\
Impressionism      & 1.0000 & 0.9870 \\
Post-Impressionism & 0.9953 & 0.9770 \\
Realism            & 0.9968 & 0.9523 \\
Renaissance        & 0.9958 & 0.9713 \\
Romanticism        & 0.9968 & 0.9658 \\
Surrealism         & 0.9948 & 0.9420 \\
Ukiyo-e            & 0.9995 & 0.9918 \\
\bottomrule
\end{tabular}
\end{adjustbox}
\end{table}

\subsection{OOD Cross-Generator Evaluation}\label{sec:ood_results}

The main question in our OOD experiment is whether detectors trained on LDM and SD2.1 artwork can still work reliably on prompt-aligned SD3.5m images that were not seen during training. Table~\ref{tab:ood_main} reports the full OOD evaluation results for all models on the~20,000-image SD3.5m test set.


\begin{table}[!htb]
\centering
\caption{OOD performance on the SD3.5m test set
(column definitions are the same as Table~\ref{tab:id_results}, and all metrics are computed at the
validation-selected threshold as per Table~\ref{tab:thresholds})}
\label{tab:ood_main}
\begin{adjustbox}{scale=0.785}
\begin{tabular}{lccccccccc}
\toprule
\textbf{Model} &
  \textbf{Bal} & \textbf{Prec} & \textbf{Rec} &
  \textbf{F1} & \textbf{FNR} & \textbf{FPR} &
  \textbf{MCC} & \textbf{AUC} & \textbf{PR} \\
\midrule
CLIP ViT-L/14   & 0.7829 & 0.9967 & 0.5676 & 0.7233 & 0.4324 & 0.0019 & 0.6268 & 0.9658 & 0.9703 \\
ConvNeXt-Base   & 0.7643 & 0.9537 & 0.5556 & 0.7021 & 0.4444 & 0.0270 & 0.5817 & 0.9256 & 0.9252 \\
EfficientNet-B0 & 0.7156 & 0.9164 & 0.4745 & 0.6252 & 0.5255 & 0.0433 & 0.4922 & 0.8799 & 0.8782 \\
ResNet-50       & 0.7125 & 0.9258 & 0.4619 & 0.6163 & 0.5381 & 0.0370 & 0.4910 & 0.8779 & 0.8811 \\
ResNet-18       & 0.6688 & 0.8369 & 0.4193 & 0.5587 & 0.5807 & 0.0817 & 0.3896 & 0.7955 & 0.7927 \\
\bottomrule
\end{tabular}
\end{adjustbox}
\end{table}

\begin{figure}[!htb]
\centering
\begin{tikzpicture}[scale=1.0, every node/.style={scale=1.0}]
\pgfkeys{/pgf/number format/.cd,1000 sep={}}
\begin{axis}[
        width  = 8.5cm,
        height = 6.0cm,
        ymin=0.0,ymax=1.025,
        ytick={0.0, 0.2, 0.4, 0.6, 0.8, 1.0},
        major x tick style = transparent,
        ybar=5*\pgflinewidth,
        bar width=25.0pt,
        ylabel = {OOD balanced accuracy},
        ylabel style = {scale = 0.9},
        symbolic x coords={A, B, C, D, E},
        xticklabels={
        		CLIP ViT-L/14,
		ConvNeXt-Base,
		EfficientNet-B0,
		ResNet-50,
		ResNet-18
		},
	y tick label style={scale=0.9,
    		/pgf/number format/.cd,
   		fixed,
   		fixed zerofill,
    		precision=2},
        xtick = data,
        x tick label style={scale=0.85,
        		rotate=60,
		anchor=north east,
		inner sep=0mm
		},
        nodes near coords,
        every node near coord/.append style={rotate=90, scale=0.785,
        								   anchor=west, 
								   /pgf/number format/.cd,
								   fixed,
								   fixed zerofill,
								   precision=3},
        enlarge x limits=0.165,
        legend cell align=left,
        legend pos=south east,
]
\addplot [fill=blue,opacity=1.00]
coordinates {
(A, 0.783)
(B, 0.764)
(C, 0.716)
(D, 0.712)
(E, 0.669)
};
\end{axis}
\end{tikzpicture}
\caption{OOD balanced accuracy across detectors}
\label{fig:ood_bar_acc}
\end{figure}
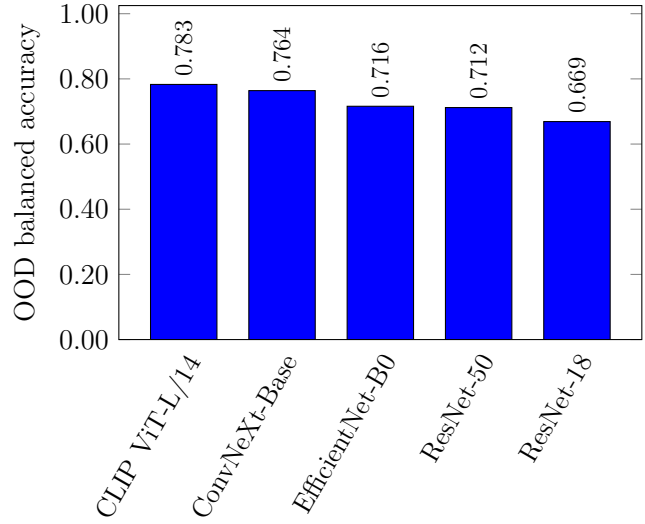
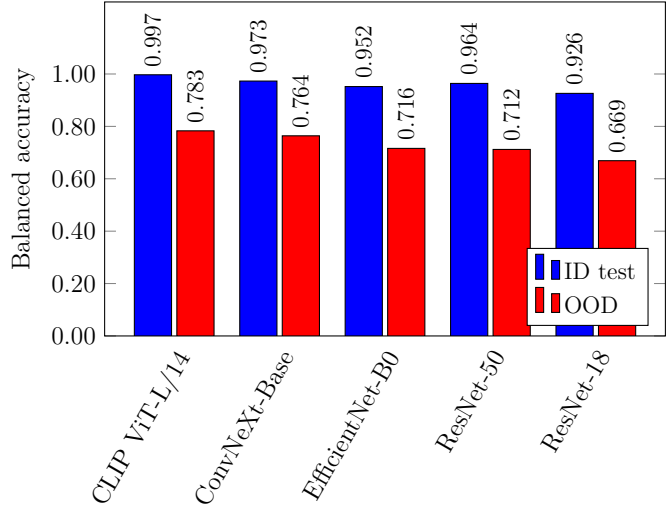
\begin{figure}[!htb]
\centering
\begin{tikzpicture}[scale=1.0, every node/.style={scale=1.0}]
\pgfkeys{/pgf/number format/.cd,1000 sep={}}
\begin{axis}[
        width  = 9.0cm,
        height = 6.0cm,
        ymin=0.0,ymax=1.275,
        ytick={0.0, 0.2, 0.4, 0.6, 0.8, 1.0},
        major x tick style = transparent,
        ybar=5*\pgflinewidth,
        bar width=14.0pt,
        ylabel = {Balanced accuracy},
        ylabel style = {scale = 0.8},
        symbolic x coords={A, B, C, D, E},
        xticklabels={
        		CLIP ViT-L/14,
		ConvNeXt-Base,
		EfficientNet-B0,
		ResNet-50,
		ResNet-18
		},
	y tick label style={scale=0.8,
    		/pgf/number format/.cd,
   		fixed,
   		fixed zerofill,
    		precision=2},
        xtick = data,
        x tick label style={scale=0.8,
        		rotate=60,
		anchor=north east,
		inner sep=0mm
		},
        nodes near coords,
        every node near coord/.append style={rotate=90, scale=0.75,
        								   anchor=west, 
								   /pgf/number format/.cd,
								   fixed,
								   fixed zerofill,
								   precision=3},
        enlarge x limits=0.165,
        legend cell align=left,
        legend pos=south east,
        legend style={nodes={scale=0.75},
        },
]
\addplot [fill=blue,opacity=1.00]
coordinates {
(A, 0.997)
(B, 0.973)
(C, 0.952)
(D, 0.964)
(E, 0.926)
};
\addlegendentry{ID test}
\addplot [fill=red,opacity=1.00]
coordinates {
(A, 0.783)
(B, 0.764)
(C, 0.716)
(D, 0.712)
(E, 0.669)
};
\addlegendentry{OOD}
\end{axis}
\end{tikzpicture}
\caption{ID-test and OOD balanced accuracy}\label{fig:ood_gap_bar}
\end{figure}

Table~\ref{tab:ood_gap} quantifies the performance drop in terms of
the absolute balanced accuracy gap ($\Delta_{\Bal}$) and absolute
recall gap ($\Delta_{\Rec}$) between the ID test set and the OOD
set. Figures~\ref{fig:ood_bar_acc} and~\ref{fig:ood_gap_bar}
visualize the OOD balanced accuracy ranking and the ID-to-OOD drop
across all models.

\begin{table}[!htb]
\centering
\caption{ID-to-OOD performance gap ($\Delta_{\text{Bal}}$ is OOD minus ID balanced accuracy; 
$\Delta_{\text{Rec}}$ is OOD minus ID recall)}\label{tab:ood_gap}
\begin{adjustbox}{scale=0.85}
\begin{tabular}{lcccc}
\toprule
\multirow{2}{*}{\raisebox{-4pt}{\textbf{Model}}} &
  \multicolumn{2}{c}{\textbf{Balanced accuracy}} 
  & \multirow{2}{*}{\raisebox{-4pt}{$\boldsymbol{\Delta_{\Bal}}$}} 
  & \multirow{2}{*}{\raisebox{-4pt}{$\boldsymbol{\Delta_{\Rec}}$}} \\ \cmidrule(lr){2-3}
&
  \ \ \ \textbf{ID} & \textbf{OOD}\ \  \\
\midrule
ResNet-18       & \ \ \ 0.9260 & 0.6688\ \  & $-$0.257 & $-$0.510 \\
ResNet-50       & \ \ \ 0.9637 & 0.7125\ \  & $-$0.251 & $-$0.501 \\
EfficientNet-B0 & \ \ \ 0.9524 & 0.7156\ \  & $-$0.237 & $-$0.474 \\
CLIP ViT-L/14   & \ \ \ 0.9969 & 0.7829\ \  & $-$0.214 & $-$0.430 \\
ConvNeXt-Base   & \ \ \ 0.9727 & 0.7643\ \  & $-$0.208 & $-$0.417 \\
\bottomrule
\end{tabular}
\end{adjustbox}
\end{table}


Every model suffers a substantial performance drop on the OOD test set. ResNet-18 experiences the
largest decrease, losing~25.7 percentage points of balanced accuracy and~51.0
percentage points of AI recall. Even the strongest model, CLIP ViT-L/14, 
loses~21.4 percentage points of balanced accuracy.

Critically, the failure mode is asymmetric. False negative rates increase sharply on the OOD set while false positive rates remain comparatively low: CLIP ViT-L/14 maintains an OOD FPR of just~0.0019, and ConvNeXt-Base maintains an OOD FPR of~0.0270. This implies that the detectors remain conservative in assigning the AI label. They rarely misclassify human paintings as AI-generated, but fail to recognize a large fraction of SD3.5m images. At the validation-selected thresholds, the five detectors miss approximately 4,300 to 5,800 of the 10,000 SD3.5m images. From a defensive perspective, these false negatives represent synthetic images that would pass the screening mechanism as human-created. In an operational workflow, such failures could allow synthetic content to bypass checks used for misinformation triage, marketplace fraud review, impersonation assessment or content-origin verification.

Among the five models, ConvNeXt-Base is the strongest CNN-based detector on the OOD test set, achieving a balanced accuracy of~0.764 and a ROC-AUC of~0.926, while CLIP ViT-L/14 performs best overall with a balanced accuracy of~0.783 and a ROC-AUC of~0.966. ResNet-50 and EfficientNet-B0 show very similar OOD performance (0.712 and 0.716 balanced accuracy, respectively), although their in-distribution results differ more noticeably. This suggests that improvements in standard in-distribution performance do not necessarily translate into proportional gains in cross-generator generalization. More broadly, the results show that strong performance on known generators may create a misleading impression of security when a deployed detector is not regularly evaluated against newly emerging generation systems~\cite{ai2023}.

\begin{figure}[!htb]
\centering
\begin{tabular}{cc}
    \adjustbox{scale=0.8}{%
\begin{tikzpicture}[scale=1.0]
    \begin{axis}[
        width=4.5cm,
        height=4.5cm,
	colormap={bluewhite}{color=(white) rgb255=(100,149,237)},
        xticklabels={
        \texttt{Human},
        \texttt{AI},
        },
        xtick={0,...,1},
        xtick style={draw=none},
	xticklabel style={font=\tt,scale=1.0},
        yticklabels={
        \texttt{Human},
        \texttt{AI},
        },
        ytick={0,...,1},
        ytick style={draw=none},
        enlargelimits=false,
        yticklabel style={xshift=0pt,font=\tt,scale=1.0},
        colorbar,
        colorbar/width=2.5mm,
        colorbar style={
            ytick={0,2500,5000,7500,10000},
            yticklabels={0,2500,5000,7500,10000},
            ytick style={draw=none},
	    scaled y ticks=false,
            yticklabel={\pgfmathprintnumber\tick},
            yticklabel style={font=\tt, scale=0.9,
            		/pgf/number format/fixed,
			/pgf/number format/fixed zerofill,
			/pgf/number format/precision=0}
        },
        point meta min=0,
        point meta max=10000,
        nodes near coords={\pgfmathprintnumber\pgfplotspointmeta},
        nodes near coords black white/.style={
            small value/.style={
                yshift=-7pt,
                text=black,
                /pgf/number format/fixed,
                /pgf/number format/precision=0,
                /pgf/number format/zerofill=true,
                scale=1.0,
            },
            large value/.style={
                yshift=-7pt,
                text=white,
                /pgf/number format/fixed,
                /pgf/number format/precision=0,
                /pgf/number format/zerofill=true,
                scale=1.0,
            },
            every node near coord/.style={
                check for zero/.code={
                    \pgfmathfloatifflags{\pgfplotspointmeta}{0}{
                        \pgfkeys{/tikz/coordinate}
                    }{
                        \begingroup
                        \pgfkeys{/pgf/fpu}
                        \pgfmathparse{\pgfplotspointmeta<#1}
                        \global\let\result=\pgfmathresult
                        \endgroup
                        %
                        %
                        \pgfmathfloatcreate{1}{1.0}{0}
                        \let\ONE=\pgfmathresult
                        \ifx\result\ONE
                            \pgfkeysalso{/pgfplots/small value}
                        \else
                            \pgfkeysalso{/pgfplots/large value}
                        \fi
                    }
                },
                check for zero,
            },
        },
        nodes near coords black white=5000,
    ]
        \addplot[
            matrix plot,
            mesh/cols=2,
            point meta=explicit,draw=gray
        ] table [meta=C] {
            x y C
0 0 9981
1 0 19
0 1 4324
1 1 5676
         };
    \end{axis}
\end{tikzpicture}
    }
    &
    \adjustbox{scale=0.8}{%
\begin{tikzpicture}[scale=1.0]
    \begin{axis}[
        width=4.5cm,
        height=4.5cm,
	colormap={bluewhite}{color=(white) rgb255=(100,149,237)},
        xticklabels={
        \texttt{Human},
        \texttt{AI},
        },
        xtick={0,...,1},
        xtick style={draw=none},
	xticklabel style={font=\tt,scale=1.0},
        yticklabels={
        \texttt{Human},
        \texttt{AI},
        },
        ytick={0,...,1},
        ytick style={draw=none},
        enlargelimits=false,
        yticklabel style={xshift=0pt,font=\tt,scale=1.0},
        colorbar,
        colorbar/width=2.5mm,
        colorbar style={
            ytick={0,2500,5000,7500,10000},
            yticklabels={0,2500,5000,7500,10000},
            ytick style={draw=none},
	    scaled y ticks=false,
            yticklabel={\pgfmathprintnumber\tick},
            yticklabel style={font=\tt, scale=0.9,
            		/pgf/number format/fixed,
			/pgf/number format/fixed zerofill,
			/pgf/number format/precision=0}
        },
        point meta min=0,
        point meta max=10000,
        nodes near coords={\pgfmathprintnumber\pgfplotspointmeta},
        nodes near coords black white/.style={
            small value/.style={
                yshift=-7pt,
                text=black,
                /pgf/number format/fixed,
                /pgf/number format/precision=0,
                /pgf/number format/zerofill=true,
                scale=1.0,
            },
            large value/.style={
                yshift=-7pt,
                text=white,
                /pgf/number format/fixed,
                /pgf/number format/precision=0,
                /pgf/number format/zerofill=true,
                scale=1.0,
            },
            every node near coord/.style={
                check for zero/.code={
                    \pgfmathfloatifflags{\pgfplotspointmeta}{0}{
                        \pgfkeys{/tikz/coordinate}
                    }{
                        \begingroup
                        \pgfkeys{/pgf/fpu}
                        \pgfmathparse{\pgfplotspointmeta<#1}
                        \global\let\result=\pgfmathresult
                        \endgroup
                        %
                        %
                        \pgfmathfloatcreate{1}{1.0}{0}
                        \let\ONE=\pgfmathresult
                        \ifx\result\ONE
                            \pgfkeysalso{/pgfplots/small value}
                        \else
                            \pgfkeysalso{/pgfplots/large value}
                        \fi
                    }
                },
                check for zero,
            },
        },
        nodes near coords black white=5000,
    ]
        \addplot[
            matrix plot,
            mesh/cols=2,
            point meta=explicit,draw=gray
        ] table [meta=C] {
            x y C
0 0 9730
1 0 270
0 1 4444
1 1 5556
         };
    \end{axis}
\end{tikzpicture}
    }
    \\
    \adjustbox{scale=0.85}{(a) CLIP ViT-L/14}
    &
    \adjustbox{scale=0.85}{(b) ConvNeXt-Base}
\end{tabular}
\caption{OOD confusion matrices}\label{fig:ood_confusion}
\end{figure}
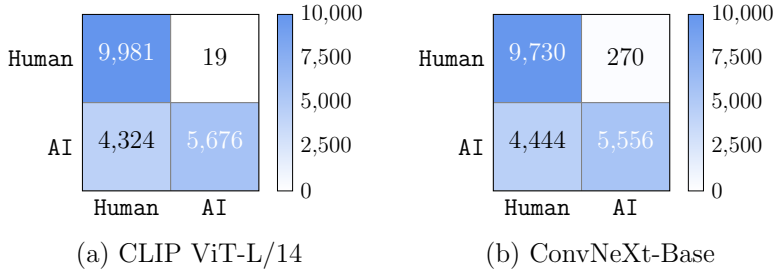

Figure~\ref{fig:ood_confusion} presents the OOD confusion matrices for CLIP ViT-L/14 and ConvNeXt-Base. The matrices show that OOD degradation is driven primarily by false negatives, with many SD3.5m AI images predicted as human, while false positives on human artwork samples remain relatively uncommon. This directly illustrates the asymmetric error pattern underlying the OOD performance drop.

\begin{figure}[!htb]
\centering
\includegraphics[scale=0.475]{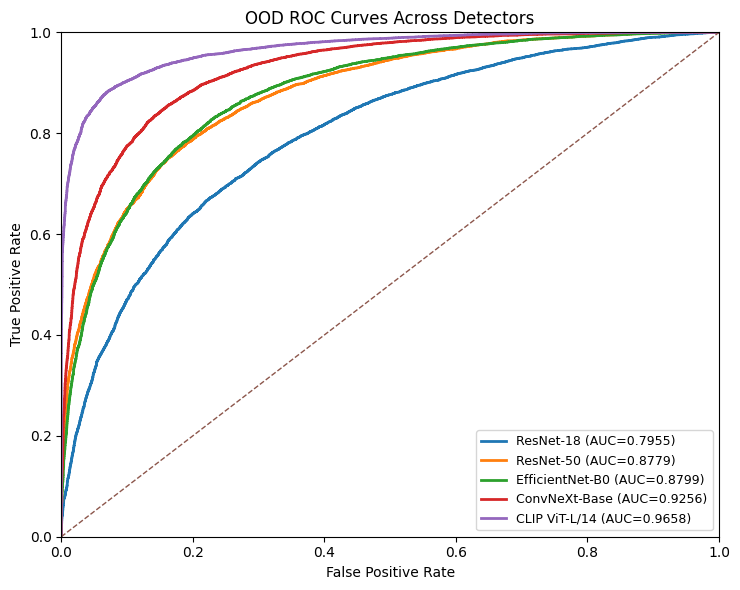}
\caption{OOD ROC curves across detectors}
\label{fig:ood_roc}
\end{figure}

Figure~\ref{fig:ood_roc} shows the OOD ROC curves for all five models overlaid.
Unlike the tightly clustered ID ROC curves in Figure~\ref{fig:id_roc}, the OOD
curves spread out visibly, reflecting the divergence in model generalization.

\begin{figure}[!htb]
\centering
\begin{tabular}{cc}
    \includegraphics[scale=0.375]{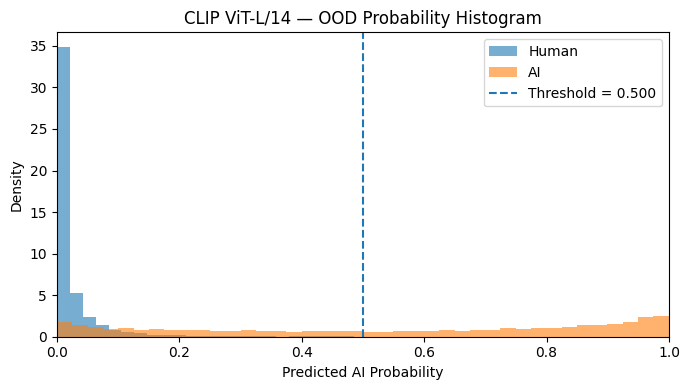}
    &
    \includegraphics[scale=0.375]{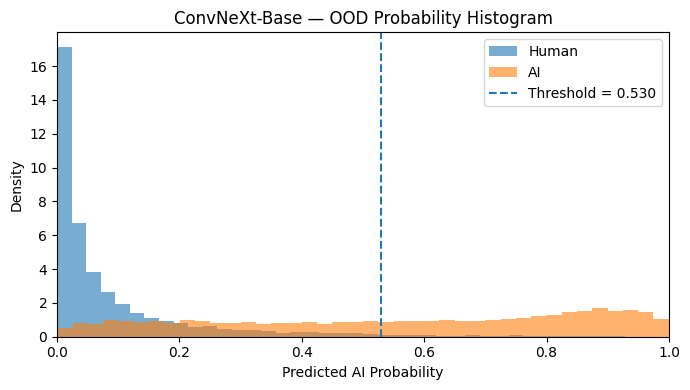}
    \\
    \adjustbox{scale=0.85}{(a) CLIP ViT-L/14}
    &
    \adjustbox{scale=0.85}{(b) ConvNeXt-Base}
\end{tabular}
\caption{OOD predicted probability histograms}\label{fig:ood_prob_hist}
\end{figure}

The probability histograms in Figure~\ref{fig:ood_prob_hist} emphasize the OOD failure mode. On the ID test set, human and AI scores are strongly separated, with human images concentrated near $p_{\AI}=0$ and AI images near $p_{\AI}=1$. In contrast, for the OOD SD3.5m dataset, this separation weakens substantially, as the AI-score distribution becomes much more spread out and shifts toward lower values, with many SD3.5m images falling below the selected threshold. As a result, a large number of OOD AI images are misclassified as human.

\begin{figure}[!htb]
\centering
\begin{tabular}{cc}
    \includegraphics[scale=0.4]{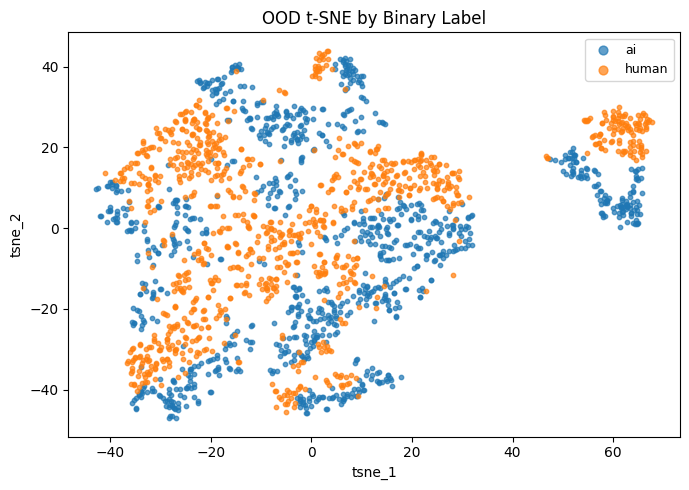}
    &
    \includegraphics[scale=0.4]{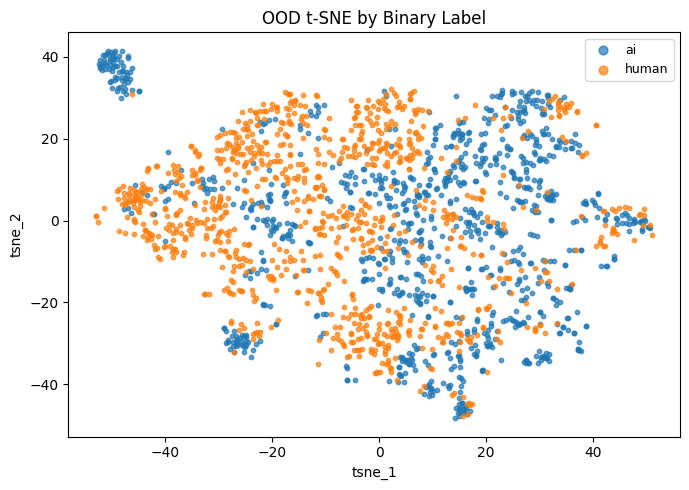}
    \\
    \adjustbox{scale=0.85}{(a) CLIP ViT-L/14}
    &
    \adjustbox{scale=0.85}{(b) ConvNeXt-Base}
\end{tabular}
\caption{t-SNE projections of the OOD feature space}\label{fig:ood_tsne}
\end{figure}

Next, we consider t-SNE
plots, which provide qualitative visualizations of the feature-space structure.
For these visualizations, a random subset of~2,000 images was sampled from
each evaluated split. Frozen backbone features were first reduced to~50 
dimensions using PCA. Then t-SNE was run with perplexity~30, PCA
initialization, and automatic learning rate selection. 
Figure~\ref{fig:ood_tsne} shows the resulting t-SNE projections of the OOD feature space
for CLIP ViT-L/14 and ConvNeXt-Base, with human and SD3.5m AI images colored
differently. The substantial overlap between the two classes suggests that the
decision boundary learned from the ID data does not transfer cleanly to the
OOD distribution.

\subsection{Style-wise OOD Performance}\label{sec:stylewise_ood}

The overall OOD metrics reported in Section~\ref{sec:ood_results}
provide a useful summary of detector performance across all ten art styles.
A style-wise analysis complements this view by showing how generator shift
manifests within individual artistic movements. Different styles have
distinct visual vocabularies: for example, Ukiyo-e is characterized by
flat color areas and bold outlines, whereas Realism emphasizes
naturalistic textures and photographic plausibility. SD3.5m may preserve
these stylistic properties to different degrees across styles. Examining
per-style OOD performance helps identify which styles remain
more robust under generator shift, which styles are more challenging, and
whether these patterns are consistent across all five backbones or vary by
architecture. 

\begin{figure}[!htb]
\centering
\begin{tikzpicture}[scale=1.0]
    \begin{axis}[
        width=10cm,
        height=7.5cm,
	colormap={redblue}{color=(blue) color=(red)},
        xticklabels={
CLIP ViT-L/14,
ConvNeXt-Base,
EfficientNet-B0,
ResNet-50,
ResNet-18
        },
        xtick={0,...,4},
        xtick style={draw=none},
	xticklabel style={anchor=east,rotate=60,yshift=-5pt,scale=0.825},
        yticklabels={
Realism,
Romanticism,
Impressionism,
Post-Impressionism,
Expressionism,
Art Nouveau,
Surrealism,
Baroque,
Renaissance,
Ukiyo\_e
        },
        ytick={0,...,9},
        ytick style={draw=none},
        enlargelimits=false,
        yticklabel style={scale=0.85},
        colorbar,
        colorbar style={
            ytick={0.6,0.7,0.8,0.9,1.0},
            yticklabels={0.6,0.7,0.8,0.9,1.0},
            yticklabel={\pgfmathprintnumber\tick},
            yticklabel style={
            		scale=0.85,
  			/pgf/number format/.cd,
  			fixed,
  			fixed zerofill,
  			precision=1}
        },
        point meta min=0.6,
        point meta max=1.0,
        nodes near coords={\pgfmathprintnumber\pgfplotspointmeta},
        nodes near coords black white/.style={
            small value/.style={
                yshift=-5pt,
                text=black,
                /pgf/number format/fixed,
                /pgf/number format/precision=3,
                /pgf/number format/zerofill=true,
                scale=0.75,
            },
            large value/.style={
                yshift=-5pt,
                text=white,
                /pgf/number format/fixed,
                /pgf/number format/precision=3,
                /pgf/number format/zerofill=true,
                scale=0.75,
            },
            every node near coord/.style={
                check for zero/.code={
                    \pgfmathfloatifflags{\pgfplotspointmeta}{0}{
                        \pgfkeys{/tikz/coordinate}
                    }{
                        \begingroup
                        \pgfkeys{/pgf/fpu}
                        \pgfmathparse{\pgfplotspointmeta<#1}
                        \global\let\result=\pgfmathresult
                        \endgroup
                        %
                        %
                        \pgfmathfloatcreate{1}{1.0}{0}
                        \let\ONE=\pgfmathresult
                        \ifx\result\ONE
                            \pgfkeysalso{/pgfplots/small value}
                        \else
                            \pgfkeysalso{/pgfplots/large value}
                        \fi
                    }
                },
                check for zero,
            },
        },
        nodes near coords black white=0.1,
    ]
        \addplot[
            matrix plot,
            mesh/cols=5,
            point meta=explicit,draw=gray
        ] table [meta=C] {
            x y C
0 0 0.694
1 0 0.665
2 0 0.661
3 0 0.647
4 0 0.630
0 1 0.740
1 1 0.702
2 1 0.674
3 1 0.673
4 1 0.636
0 2 0.752
1 2 0.708
2 2 0.685
3 2 0.693
4 2 0.658
0 3 0.754
1 3 0.743
2 3 0.680
3 3 0.684
4 3 0.655
0 4 0.750
1 4 0.746
2 4 0.707
3 4 0.678
4 4 0.654
0 5 0.709
1 5 0.839
2 5 0.717
3 5 0.722
4 5 0.692
0 6 0.837
1 6 0.782
2 6 0.725
3 6 0.687
4 6 0.654
0 7 0.833
1 7 0.738
2 7 0.715
3 7 0.736
4 7 0.668
0 8 0.794
1 8 0.776
2 8 0.726
3 8 0.742
4 8 0.708
0 9 0.966
1 9 0.943
2 9 0.865
3 9 0.861
4 9 0.732
         };
    \end{axis}
\end{tikzpicture}
\caption{Heatmap of OOD balanced accuracy across art styles and detectors}\label{fig:ood_style_heatmap}
\end{figure}
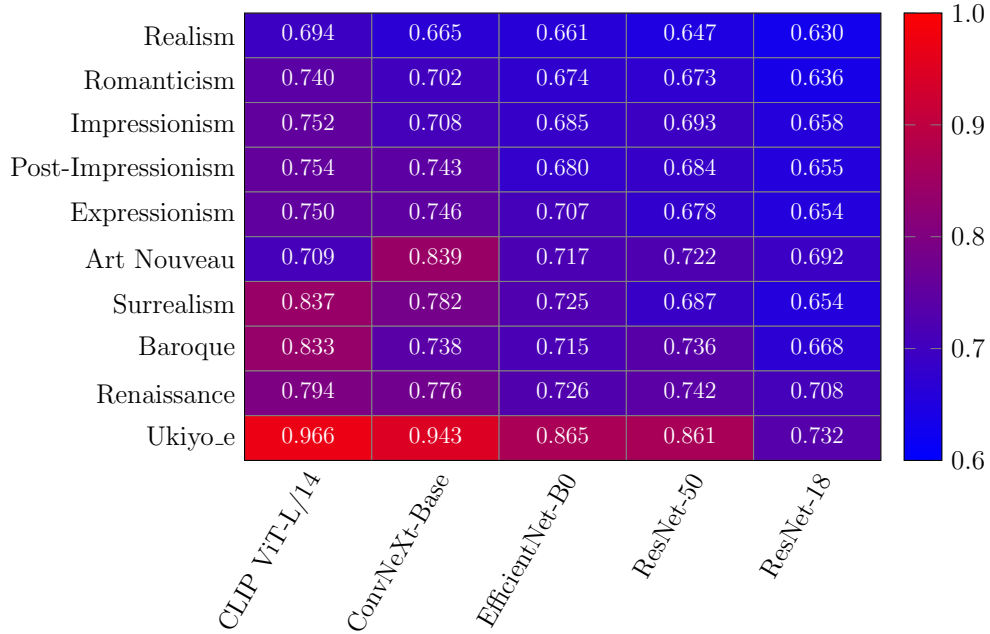

Table~\ref{tab:ood_stylewise_bal} reports style-wise OOD balanced accuracy
for all five detectors. Each cell corresponds to the~2,000-image OOD subset
for that style (1,000 human and~1,000 SD3.5m AI), and the rightmost column
gives the mean across all five models. Figure~\ref{fig:ood_style_heatmap}
visualizes the same style-by-model matrix as a heatmap, making the overall
difficulty pattern across detector backbones easier to compare. 
Ukiyo-e (bottom row) is consistently the easiest style for all
models, whereas Realism (top row) is the hardest, with no model
exceeding a balanced accuracy of~0.694 for Realism.

\begin{table}[!htb]
\centering
\caption{OOD balanced accuracy per style for all five detectors}\label{tab:ood_stylewise_bal}
\begin{adjustbox}{scale=0.85}
\begin{tabular}{lcccccc}
\toprule
\textbf{Style} &
  \textbf{RN18} & \textbf{RN50} & \textbf{ENB0} &
  \textbf{CNX} & \textbf{CLIP} & \textbf{Mean} \\
\midrule
Realism            & 0.631 & 0.647 & 0.661 & 0.666 & 0.694 & 0.660 \\
Romanticism        & 0.636 & 0.673 & 0.675 & 0.702 & 0.740 & 0.685 \\
Impressionism      & 0.659 & 0.693 & 0.686 & 0.708 & 0.752 & 0.700 \\
Post-Impressionism & 0.655 & 0.684 & 0.681 & 0.743 & 0.754 & 0.703 \\
Expressionism      & 0.655 & 0.678 & 0.707 & 0.746 & 0.751 & 0.707 \\
Art Nouveau        & 0.692 & 0.723 & 0.717 & 0.840 & 0.709 & 0.736 \\
Surrealism         & 0.654 & 0.687 & 0.725 & 0.782 & 0.837 & 0.737 \\
Baroque            & 0.668 & 0.737 & 0.715 & 0.738 & 0.833 & 0.738 \\
Renaissance        & 0.708 & 0.743 & 0.726 & 0.776 & 0.795 & 0.750 \\
Ukiyo-e            & 0.733 & 0.862 & 0.865 & 0.943 & 0.966 & 0.874 \\
\midrule
\textbf{Average} 
	& \textbf{0.669} & \textbf{0.712} & \textbf{0.716} & \textbf{0.764} & \textbf{0.783} & \textbf{0.729} \\
\bottomrule
\end{tabular}
\end{adjustbox}
\end{table}

\begin{figure}[!htb]
\centering
\begin{tikzpicture}[scale=1.0, every node/.style={scale=1.0}]
\pgfkeys{/pgf/number format/.cd,1000 sep={}}
\begin{axis}[
        width  = 8.5cm,
        height = 6.0cm,
        ymin=0.0,ymax=1.0,
        ytick={0.0, 0.2, 0.4, 0.6, 0.8, 1.0},
        major x tick style = transparent,
        ybar=5*\pgflinewidth,
        bar width=14.0pt,
        ylabel = {Mean OOD recall},
        ylabel style = {scale = 0.8},
        symbolic x coords={A, B, C, D, E, F, G, H, I, J},
        xticklabels={
        		Realism,
		Romanticism,
		Impressionism,
		Post-Impressionism,
		Expressionism,
		Art Nouveau,
		Surrealism,
		Baroque,
		Renaissance,
		Ukiyo\_e
		},
	y tick label style={scale=0.8,
    		/pgf/number format/.cd,
   		fixed,
   		fixed zerofill,
    		precision=2},
        xtick = data,
        x tick label style={scale=0.8,
        		rotate=60,
		anchor=north east,
		inner sep=0mm
		},
        nodes near coords,
        every node near coord/.append style={rotate=90, scale=0.75,
        								   anchor=west, 
								   /pgf/number format/.cd,
								   fixed,
								   fixed zerofill,
								   precision=3},
        enlarge x limits=0.085,
        legend cell align=left,
        legend pos=south east,
]
\addplot [fill=blue,opacity=1.00]
coordinates {
(A, 0.337)
(B, 0.403)
(C, 0.427)
(D, 0.436)
(E, 0.474)
(F, 0.509)
(G, 0.510)
(H, 0.535)
(I, 0.568)
(J, 0.758)
};
\end{axis}
\end{tikzpicture}
\caption{Mean OOD AI recall across all detectors per style}\label{fig:style_mean_recall_bar}
\end{figure}
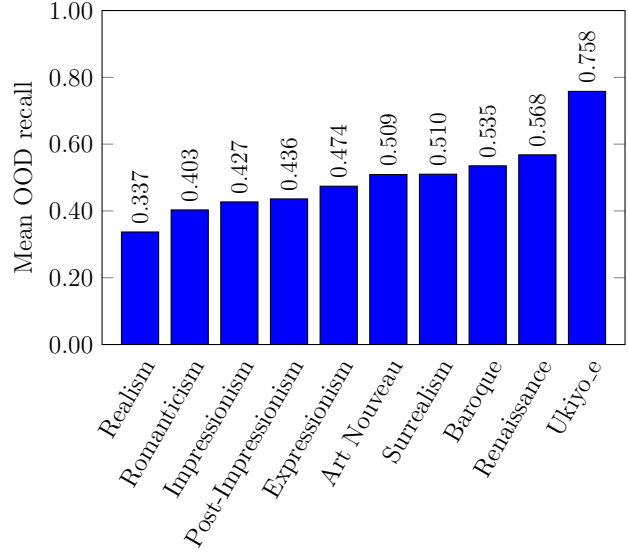

Recall is particularly important in the OOD setting because it measures the
detector's ability to correctly identify SD3.5m AI images as AI. Style-wise, recall provides a useful complement to balanced
accuracy by showing which styles most often lead to missed AI detections. 
Figure~\ref{fig:style_mean_recall_bar} presents this recall-based
view and highlights which styles are prone to false negatives under generator shift.



The styles are ordered from hardest to easiest by mean OOD AI recall, giving a model-agnostic view of style-level difficulty under generator shift. Realism is the hardest style at~0.337. Ukiyo-e is the easiest at~0.758, while all other styles fall below~0.57.
Ukiyo-e stands out because its visual structure (flat color regions, bold outlines and clear linework) remains distinctive in the SD3.5m output, making these images easier for detectors to separate from human artwork samples. Realism is the most difficult style as the detectors miss about two thirds of SD3.5m images in this category. Romanticism, Impressionism, and Post-Impressionism form the next hardest group, each with recall below~0.44, suggesting that SD3.5m images in these styles are more visually similar to authentic human paintings in color and composition.

\subsection{Source-wise In-Distribution Performance}\label{sec:sourcewise_id}

The in-distribution data consists of images from three source classes:
human artwork samples, LDM-generated images, and SD2.1-generated images. Although
the binary task merges LDM and SD2.1 into a single AI class, examining
performance by source reveals whether detectors treat both generators
similarly or whether one is easier to identify than the other. This also
provides a useful baseline for understanding how each model distributes
its errors between false positives on human images and false negatives on
each AI generator, which in turn helps contextualize the OOD results.

Table~\ref{tab:sourcewise_id} reports source-wise performance for all
models. These results show how errors are distributed between false positives on
human artwork samples and false negatives for both AI generators.

\begin{table}[!htb]
\centering
\caption{Source-wise performance on ID test set}\label{tab:sourcewise_id}
\begin{adjustbox}{scale=0.85}
\begin{tabular}{lccc}
\toprule
\textbf{Model} &
  \textbf{FPR$_\text{human}$} &
  \textbf{Recall$_\text{LDM}$} &
  \textbf{Recall$_\text{SD2.1}$} \\
\midrule
CLIP ViT-L/14   & 0.003\zz\zz\zz & 0.995\zz & 0.999\zz\zz \\
ConvNeXt-Base   & 0.027\zz\zz\zz & 0.960\zz & 0.985\zz\zz \\
ResNet-50       & 0.035\zz\zz\zz & 0.949\zz & 0.976\zz\zz \\
EfficientNet-B0 & 0.044\zz\zz\zz & 0.928\zz & 0.968\zz\zz \\
ResNet-18       & 0.077\zz\zz\zz & 0.918\zz & 0.939\zz\zz \\
\bottomrule
\end{tabular}
\end{adjustbox}
\end{table}

Across all five models, SD2.1 images are consistently detected at a
higher rate than LDM images, with SD2.1 recall exceeding LDM recall by
up to four percentage points, depending on the backbone. One possible
explanation is that SD2.1, with its default sampling configuration,
introduces slightly more distinctive statistical patterns at the
resolutions used here, though a definitive attribution would require
further investigation. The distribution of errors on human images also varies with model capacity: ResNet-18 produces a human false positive rate of~0.077, whereas CLIP ViT-L/14 reduces this to~0.003, consistent with higher-capacity backbones learning more selective decision boundaries. On the ID test set, every model achieves recall above~0.91 for both LDM
and SD2.1, whereas on the OOD set, every model falls below~0.57 recall on
SD3.5m images. This contrast suggests that the performance drop
is linked to the specific characteristics of the new generator rather
than a general degradation of detector performance.

\begin{figure}[!htb]
\centering
\begin{tabular}{cc}
    \centering
    \includegraphics[scale=0.385]{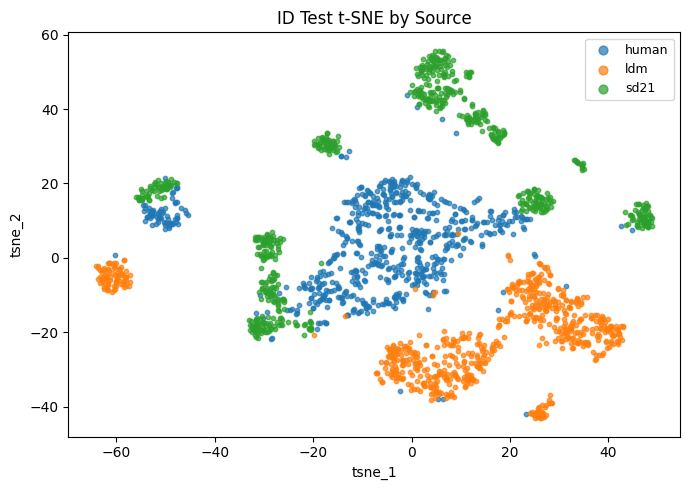}
    &
    \includegraphics[scale=0.385]{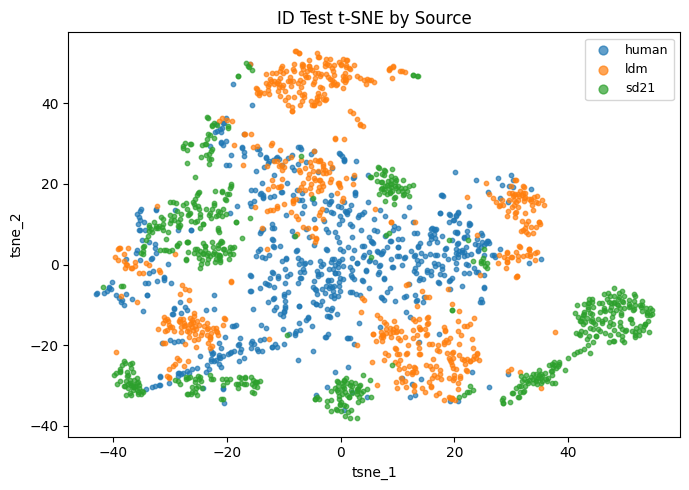}
    \\
    \adjustbox{scale=0.85}{(a) CLIP ViT-L/14}
    &
    \adjustbox{scale=0.85}{(b) ConvNeXt-Base}
\end{tabular}
\caption{t-SNE projections of ID test feature space colored by source class}\label{fig:id_tsne_source}
\end{figure}

Figure~\ref{fig:id_tsne_source} shows the ID test t-SNE colored by source
class. These results show that LDM and SD2.1 images form distinguishable
sub-clusters within the broader AI region.

\subsection{Explainability and Failure Analysis}\label{sec:analysis}

Previous sections measured the OOD performance drop caused by generator
shift. This section adds a qualitative perspective by examining where the
detector focuses when making correct and incorrect predictions. We apply Grad-CAM to ConvNeXt-Base, which is the best performing of the CNN-based detectors considered in this study. The final feature map of the ConvNeXt-Base backbone was used as the target layer and the AI class was used as the target output~\cite{selvaraju2017gradcam}. Since the classifier produces a single AI logit \(z\), the human-target attribution was computed using the complementary score \(-z\), while the AI-target attribution was computed using \(z\).

Three cases are examined including correctly detected ID AI images, OOD false negatives, and OOD false positives. These attribution maps help show which image regions support successful detection and how the model's focus changes when it fails under generator shift. The Grad-CAM maps are used only as qualitative visualizations of model behavior and not as complete explanations of the model's decision process.

 \subsubsection{Detection Success: ID True Positives}\label{sec:gradcam_tp}

\begin{figure}[!htb]
\centering
\includegraphics[width=0.85\textwidth]{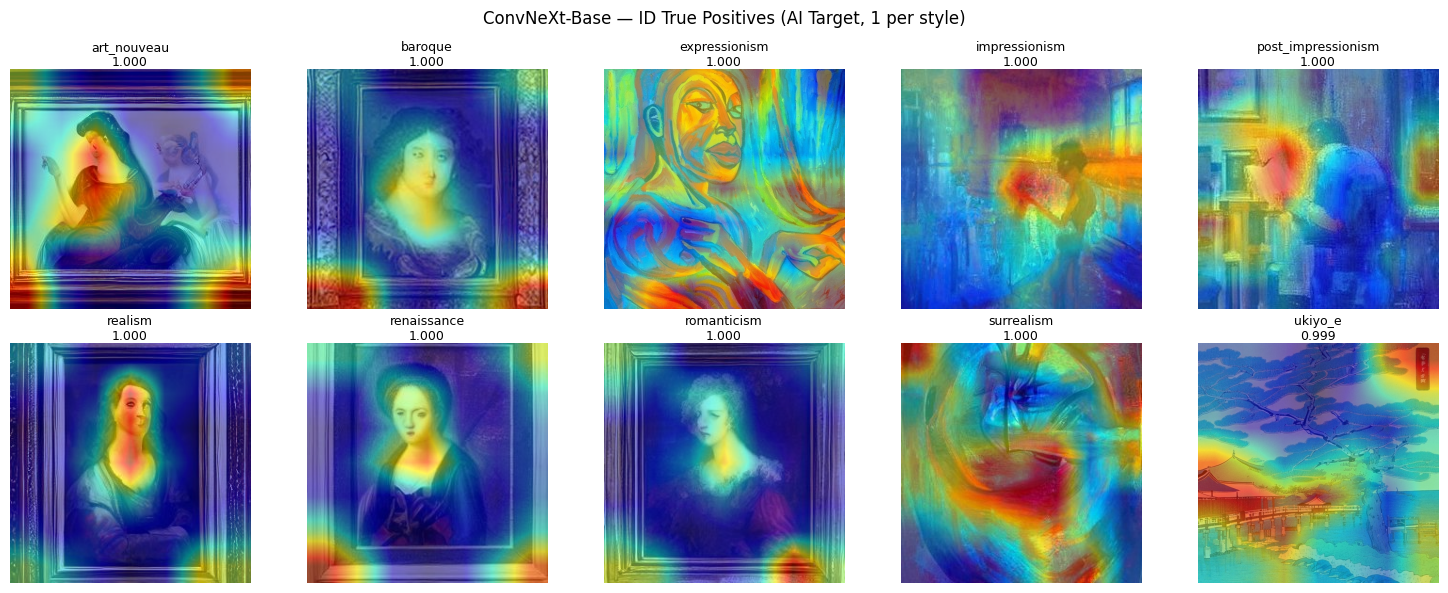}
\caption{Grad-CAM attribution maps for ID true positives with ConvNeXt-Base}
\label{fig:gradcam_id_tp}
\end{figure}

Figure~\ref{fig:gradcam_id_tp} shows Grad-CAM overlays for one confidently
detected AI (LDM or SD2.1) image per style. Specifically, the image selected is that which is assigned the highest
AI probability for each style on the ID test set.

For ID true positives, the activation maps are usually focused on specific parts of the image rather than spread across the full canvas. In portrait-like styles such as Baroque, Realism, Renaissance and Romanticism, the strongest responses appear around faces, neck regions, upper body outlines and sometimes near frame boundaries. In Expressionism, Impressionism and Post-Impressionism, the activation is concentrated near the main subject and nearby high contrast or strongly textured regions. Surrealism and Ukiyo-e show a similar pattern, but the activated regions are spread across distinctive parts of the composition rather than one small object. Overall, these maps suggest that the detector uses local visual cues to identify in-distribution AI-generated images instead of looking at the image holistically.

\subsubsection{Detection Fails: OOD False Negatives}\label{sec:gradcam_fn}

\begin{figure}[!htb]
\centering
\includegraphics[width=0.85\textwidth]{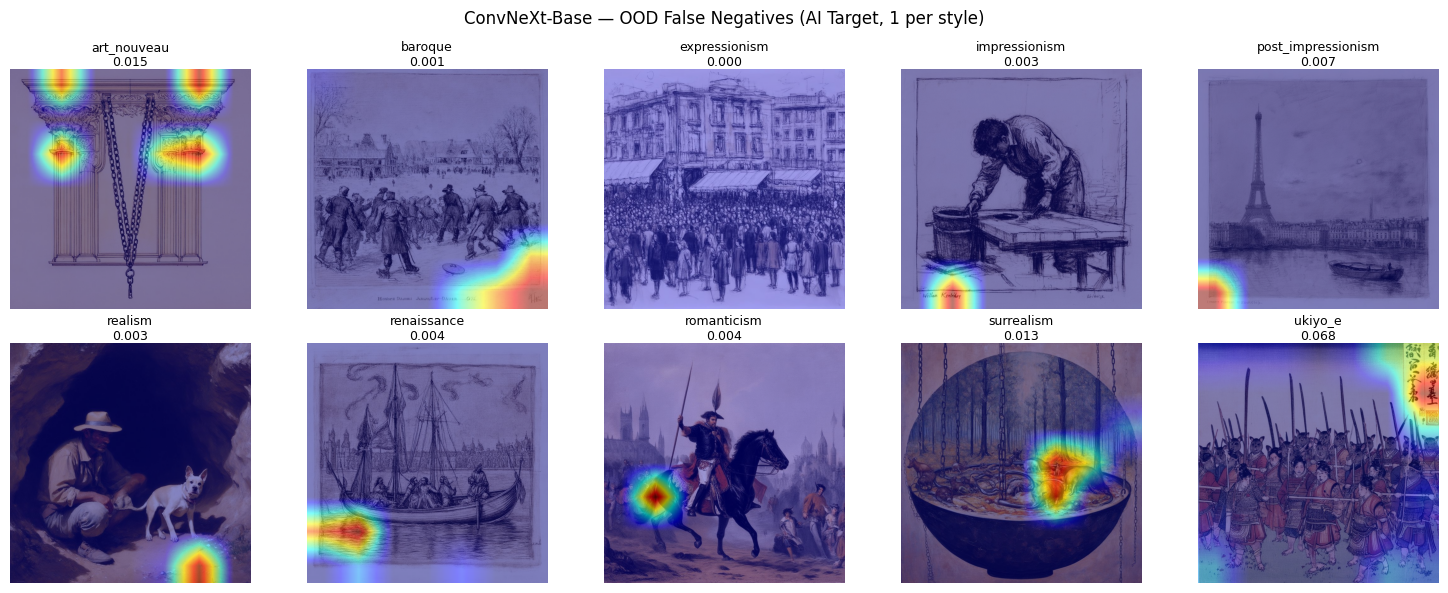}
\caption{Grad-CAM attribution maps for OOD false negatives with ConvNeXt-Base}
\label{fig:gradcam_ood_fn}
\end{figure}

Figure~\ref{fig:gradcam_ood_fn} shows Grad-CAM overlays for the OOD false
negatives, one SD3.5m image per style selected as the most convincingly
human-like miss (i.e., lowest predicted AI probability) for each style. The contrast between Figures~\ref{fig:gradcam_id_tp} and~\ref{fig:gradcam_ood_fn}
offers a visual perspective on the difference between successful
ID detection and OOD failure.

For OOD false negatives, the Grad-CAM maps are weaker and less clear than the maps for correct ID AI detections. The model does not focus strongly on the main subject or other important parts of the image. Instead, it looks at small areas near the image edges, frame-like regions or isolated texture details. In some cases, the activation appears in only one or two small spots. These weak and peripheral attribution patterns are associated with a lower predicted AI scores. These observations are consistent with the OOD results and indicate that SD3.5m does not produce the same AI-related cues that the detector learned from LDM and SD2.1 images.

\begin{figure}[!htb]
\centering
\includegraphics[width=0.8\textwidth]{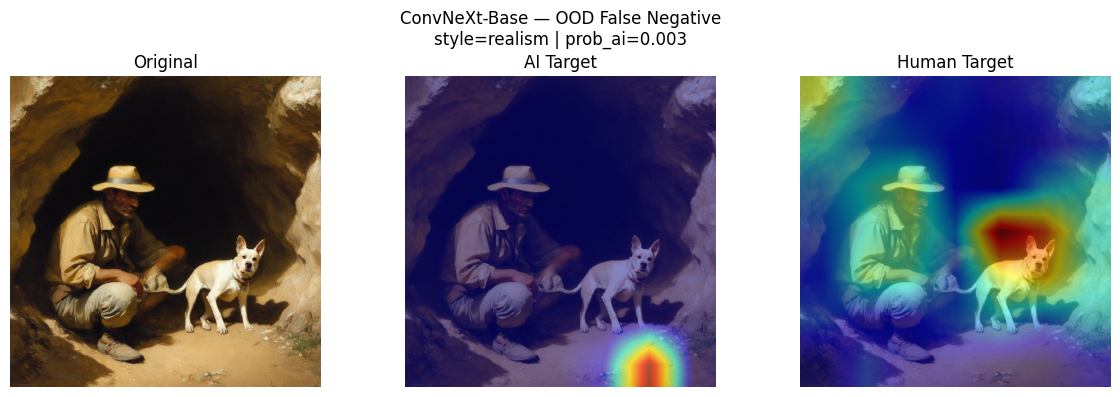}
\caption{Grad-CAM for an OOD false negative sample with ConvNeXt-Base}\label{fig:gradcam_fn_comparison}
\end{figure}

Figure~\ref{fig:gradcam_fn_comparison} shows a representative OOD false negative analyzed with both AI-target and human-target Grad-CAM. In this example, the AI-target map is weak and concentrated near a peripheral lower-image region, while the human-target map is stronger and more focused on the dog and surrounding foreground. This suggests that the highlighted regions provide stronger evidence against the AI classification than evidence supporting it, pushing the model toward a human prediction.

\subsubsection{False Alarms: OOD False Positives}\label{sec:gradcam_fp}

\begin{figure}[!htb]
\centering
\includegraphics[width=0.85\textwidth]{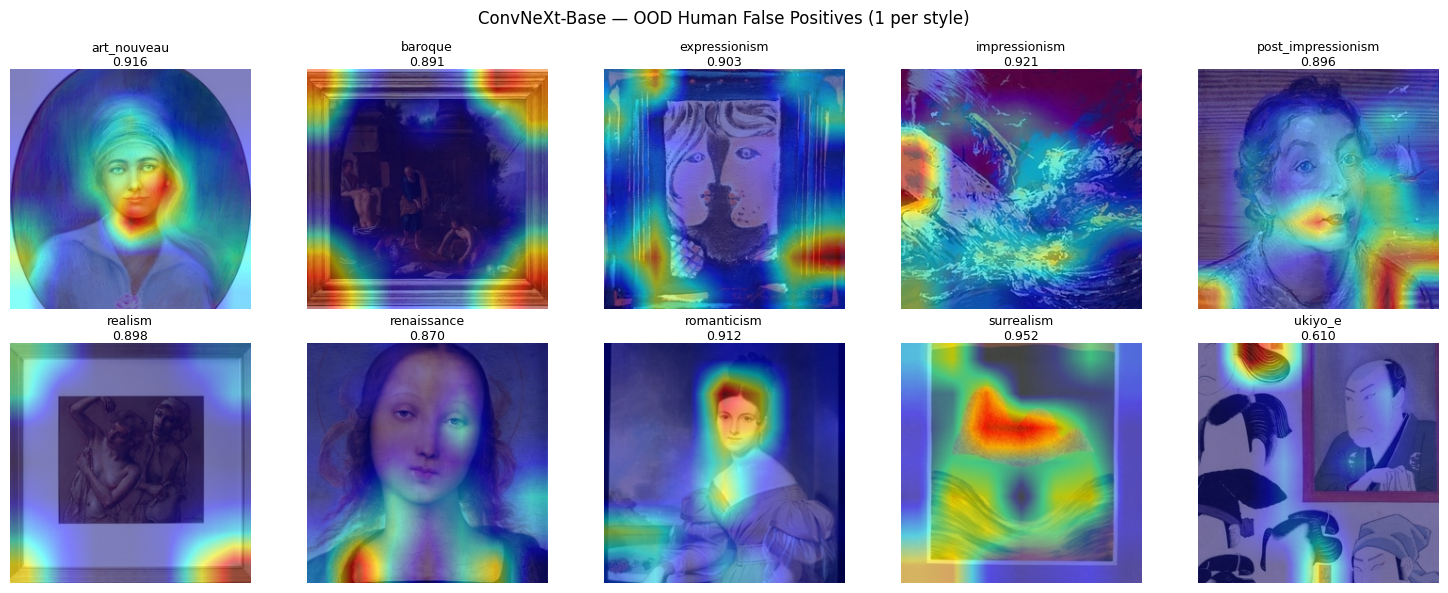}
\caption{Grad-CAM attribution maps for OOD false positives with ConvNeXt-Base}
\label{fig:gradcam_ood_fp}
\end{figure}

Figure~\ref{fig:gradcam_ood_fp} shows Grad-CAM overlays for OOD false positives. It includes one human artwork from each style that was incorrectly classified as AI. For each style, the selected example is the human artwork with the highest predicted AI probability.

For OOD false positives, the Grad-CAM maps appear more organized than the maps for OOD false negatives. In many cases, the model focuses on faces, facial outlines, frame boundaries or other clear local regions instead of spreading attention across the whole image. This suggests that false positives are caused by specific visual patterns that the detector connects with the AI class, although the image is actually human-made. These mistakes are not very common, since the overall OOD false positive rate is low, at roughly~0.2\%\ to~8.2\%\ across models. But when these errors do occur, the Grad-CAM maps show that the model is usually reacting to certain parts of the image rather than the full image as a whole.

\subsubsection{Qualitative Summary}\label{sec:analysis_discussion}

Overall, the Grad-CAM results provide qualitative evidence consistent with the OOD performance drop. For in-distribution images, ConvNeXt-Base focuses on clear local regions that appear useful for detecting LDM and SD2.1 images. These cues allow the linear classifier to separate AI-generated images from human artwork samples.

For the SD3.5m OOD set, these cues are weaker and less consistent. Many false negatives show only small or poorly placed activation regions that suggest the detector did not find the same AI-related evidence in SD3.5m images. This supports the quantitative results, where the main OOD failure comes from missed AI detections.

The style-wise patterns also match the Grad-CAM results. Realism and Romanticism show weaker activation which is consistent with their lower OOD recall. In contrast, Ukiyo-e shows more structured activation around distinctive visual elements such as flat color regions, bold outlines and calligraphic markings. This is consistent with Ukiyo-e being the easiest OOD style.

CLIP ViT-L/14 shows the same general failure pattern with low false positives but many missed SD3.5m images. Nevertheless, it performs the best overall, suggesting that its broader semantic features are more robust under generator shift.

\section{Conclusion}\label{chap:conclusion}

In this chapter, we focused on whether AI-art detectors trained on earlier diffusion
generators remain viable when applied to images produced by a newer,
architecturally different model. To study this question, a prompt-aligned
Stable Diffusion~3.5 Medium (SD3.5m) dataset was constructed across~10 art
styles and used for out-of-distribution evaluation alongside a held-out
human reference set. Five frozen-backbone detectors, namely, ResNet-18,
ResNet-50, \hbox{EfficientNet-B0}, ConvNeXt-Base, and CLIP ViT-L/14, were then
trained on in-distribution artwork from LDM and SD2.1 and evaluated under a common experimental setup.

The results reveal a consistent pattern. All five detectors achieve
strong in-distribution performance, indicating that the frozen-backbone,
linear-probe design is effective when training and test data come from the
same generator family. However, under generator shift to SD3.5m, every model
experiences a substantial drop in performance. The degradation is strongly
asymmetric with false positive rates on human artwork samples remaining low while recall on SD3.5m images falls sharply. 
This means the primary failure is not an increase in false alarms on human artwork, but a substantial increase in missed detections of images produced by the newer generator. Among the five models, CLIP
ViT-L/14 achieves the strongest OOD performance overall, while
ConvNeXt-Base is the best performing CNN-based model. Style-wise analysis further shows
that generator shift does not affect all artistic movements equally, with
Ukiyo-e remaining comparatively easy and Realism being the most
difficult style under OOD evaluation.

The qualitative analysis also supports these findings. Grad-CAM maps for
in-distribution true positives show more focused and structured activation,
whereas OOD false negatives often produce weaker and more fragmented
responses. This suggests that the visual cues learned from LDM and SD2.1 do not transfer cleanly to SD3.5m. Overall, the results indicate
that strong in-distribution accuracy is not sufficient evidence of detector
robustness when the underlying generator changes. More broadly, they highlight a practical limitation of current AI-art detectors: systems that
perform well on known generators may generalize poorly to newer models
with different synthesis mechanisms. 

This limitation has direct cybersecurity implications and demonstrates why AI-art detectors should not be treated as static or standalone mechanisms for content-authenticity verification. A detector that performs almost perfectly on known generators may still allow a substantial fraction of images from an emerging generator to pass as human-created. Such false negatives could weaken screening workflows that support misinformation review, the investigation of fraudulent authenticity claims, impersonation assessment, and content-origin verification. The risk is particularly important because it does not require a sophisticated or adaptive adversary. Simply switching to a newer generator may be sufficient to reduce detector reliability, creating a low barrier evasion pathway that does not require direct access or manipulation of the detector. Image-based detection should therefore be used as one signal within a layered defensive process that also considers provenance metadata, watermark or content-credential verification, source analysis, drift monitoring, periodic evaluation against newly released generators and human review. 



The findings of this study point to several directions for future work. First, the detectors are evaluated in a frozen-backbone, linear-probe setting, which enables controlled comparison across architectures. However, future work could test whether backbone fine-tuning or parameter-efficient adaptation can improve cross-generator robustness. Second, the OOD evaluation uses only Stable Diffusion 3.5 Medium as the new generator, so testing across additional recent image generation models would help determine whether the same failure pattern appears more broadly. Also, even though the SD3.5m dataset is prompt aligned with human artworks, there may still be some prompt distribution shift. This is because the SD3.5m images are generated using reverse prompted and title augmented prompts rather than the original AI-ArtBench prompt pipeline. Third, all detectors operate on resized inputs, which may suppress fine-scale forensic cues available at higher resolutions, suggesting that multi-scale or frequency-aware approaches could be valuable extensions. At the same time, extending the framework from binary human-versus-AI detection to multi-generator attribution would provide a more detailed view of how different synthesis pipelines leave detectable traces.


Overall, the prompt-aligned dataset construction and OOD evaluation performed in this chapter provide a practical foundation for future studies and broader research on robustness in AI-art detection and its application in cybersecurity workflows.

\bibliographystyle{plain}
\bibliography{references}

%

\end{document}